\documentclass{article}

\usepackage{arxiv}

\usepackage[utf8]{inputenc} 
\usepackage[T1]{fontenc}    
\usepackage{hyperref}       
\usepackage{url}            
\usepackage{booktabs}       
\usepackage{amsfonts}       
\usepackage{nicefrac}       
\usepackage{microtype}      
\usepackage{lipsum}
\usepackage{graphicx}
\usepackage{cite}
\usepackage{amsmath,amssymb,amsfonts}
\usepackage{multirow}
\usepackage{graphicx}
\usepackage{textcomp}
\usepackage{xcolor}
\usepackage{algorithm}
\usepackage{algpseudocode}
\usepackage{booktabs}
\usepackage{natbib}   
\usepackage{float}
\usepackage{stfloats}
\usepackage{caption}
\usepackage{subcaption}
\usepackage{placeins}

\graphicspath{ {./images/} }

\title{Deep Reinforcement Learning for Reliability Based Bi-Objective Portfolio Optimization}

\author{
Sounaq Das \\
Indian Institute of Management Amritsar \\
Amritsar, India \\
\texttt{dassounaq@gmail.com} \\
\And
Tanmay Sen \\
SQC \& OR Unit\\
Indian Statistical Institute Kolkata \\
Kolkata, India \\
\texttt{tanmay.sen@isical.ac.in} \\
\And
Raghu Nandan Sengupta \\
Department of Management Sciences \\
Indian Institute of Technology Kanpur \\
Kanpur – 208 016, India \\
\texttt{raghus@iitk.ac.in } \\
\And
Aditya Gupta \\
McKinsey and Company \\
Gurgaon, India \\
\texttt{ 
aditya\_gupta-guva@mckinsey.com } \\
}

\begin{document}
\maketitle

\begin{abstract}



Portfolio optimization under uncertainty is inherently a multi-objective decision problem involving complex interactions among return, risk, market dynamics, and practical investment constraints. Existing reliability based portfolio optimization approaches primarily rely on static optimization frameworks and often fail to capture sequential decision making, tail risk, and market frictions such as transaction costs. To address these limitations, we propose a deep reinforcement learning framework for multi-objective reliability based portfolio optimization (MORP-DRL). The proposed framework jointly optimizes expected return and downside risk using three complementary risk measures: variance, Conditional Value-at-Risk (CVaR), and Entropic Value-at-Risk (EVaR). To model uncertainty and heavy-tailed market behavior, asset returns are represented using GARCH(1,1), Extreme Value Theory, and a \(t\)-copula dependence structure, while realistic scenarios are generated through quasi-Monte Carlo simulation. A Proximal Policy Optimization (PPO) based strategy is developed under practical constraints including transaction costs and portfolio bounds, and is benchmarked against NSGA-II. Experiments on ten global equity indices across pre-COVID, COVID, and post-COVID market regimes demonstrate that MORP-DRL achieves competitive risk-return performance, reduced downside risk during periods of market stress, and scalability to high-dimensional portfolio settings.

\end{abstract}


\section{Introduction}

Portfolio optimization is one of the most fundamental problems in quantitative finance and investment management. The primary objective is to allocate capital among different financial assets in a way that balances expected return and investment risk while satisfying practical investment constraints. The seminal work of Markowitz \cite{markowitz1952, fabozzi2008portfolio} introduced the modern portfolio theory (MPT), which formulated portfolio selection as a mean-variance optimization problem. This framework established the foundation of modern portfolio management by emphasizing diversification and the trade-off between return and risk. Subsequent developments, including the capital asset Pricing model (CAPM) \cite{fama2004capital}, Black-Litterman model, and factor based investment strategies, further extended classical portfolio theory by incorporating systematic market risk and investor preferences. Despite their importance, these traditional models rely on several simplifying assumptions, such as normally distributed returns, linear dependence structures, constant covariance matrices, and frictionless markets. However, financial markets behave very differently. Asset returns frequently exhibit sudden fluctuations, changing dependence patterns, periods of high volatility, and extreme market movements, particularly during financial crises and uncertain
economic conditions. As a result, traditional variance-based approaches may not adequately capture downside and extreme market risks. To address these limitations, alternative downside risk measures have been extensively investigated in the literature. Value-at-Risk (VaR) \cite{linsmeier2000value} emerged as one of the earliest and most widely adopted downside risk measures in financial risk management. Later, \cite{ROCKAFELLAR20021443} introduced Conditional Value-at-Risk (CVaR), which measures the expected loss beyond the VaR threshold and satisfies the properties of a coherent risk measure. More recently, Entropic Value-at-Risk (EVaR) \cite{RAMOS2023120412, RIGHI2018105} and related tail-risk measures have gained attention due to their stronger theoretical properties and ability to capture extreme losses under non-Gaussian market conditions. 

At the same time, practical portfolio optimization problems have become increasingly challenging due to the incorporation of realistic investment constraints such as transaction costs, cardinality restrictions, minimum and maximum holding limits, and budget constraints. The inclusion of such constraints often leads to highly nonlinear and non-convex optimization problems that are difficult to solve using conventional convex optimization techniques. To overcome these challenges, several studies have explored metaheuristic and evolutionary optimization methods, including Genetic Algorithms (GA), Particle Swarm Optimization (PSO), Ant Colony Optimization (ACO), and Non-dominated Sorting Genetic Algorithm-II (NSGA-II) \cite{article1, SALO20241, ERTENLICE201836}. These approaches have demonstrated strong capabilities in handling multi-objective optimization problems and complex feasible regions.

In recent years, the rapid growth of financial data and increasing market complexity have encouraged the use of Machine Learning (ML) and Deep Learning (DL)\cite{goodfellow2016deep} techniques in portfolio optimization \cite{fischer2018deep, behera2023prediction}. Early ML based approaches primarily focused on predicting asset returns using methods such as Support Vector Machines, Random Forests, and Artificial Neural Networks. Although these approaches improved predictive capability, portfolio decisions were typically generated through separate static optimization procedures and lacked adaptability to rapidly changing market conditions. Furthermore, these methods often failed to capture the sequential and dynamic nature of investment decisions. To address these limitations, Deep Reinforcement Learning (DRL) \cite{arulkumaran2017deep, choudhary2025risk} has emerged as a promising paradigm for dynamic portfolio management by learning allocation strategies directly through interactions with market environments.  Unlike supervised learning methods, DRL can optimize long term investment objectives while continuously adapting to evolving market conditions. Recent studies have shown the effectiveness of actor-critic architectures, Deep Q-Networks, and policy gradient methods in financial applications \cite{gu2020empirical}.  Despite recent progress, several important limitations remain insufficiently addressed in the existing literature. First, many DRL based portfolio optimization frameworks either neglect transaction costs or model them in a simplified manner, resulting in unrealistic rebalancing strategies. Second, although tail-risk measures such as CVaR and EVaR have been investigated independently \cite{choudhary2026cvar}, their integration within a reliability based portfolio optimization framework remains limited. Third, most existing DRL approaches do not explicitly incorporate probabilistic reliability constraints to ensure robust portfolio feasibility under uncertain market conditions and nonlinear dependence structures. Finally, comparative studies between classical evolutionary optimization approaches and DRL based optimization across different market regimes remain relatively scarce.


Motivated by these research gaps, this study proposes a reliability based Bi-objective portfolio optimization framework integrated with Deep Reinforcement Learning. The proposed framework incorporates  transaction costs, probabilistic reliability constraints, and multiple downside risk measures within a unified optimization setting. In particular, three complementary portfolio optimization models are developed using variance, Conditional Value-at-Risk (CVaR), and Entropic Value-at-Risk (EVaR) as alternative risk measures. To capture nonlinear dependence and tail co-movement among financial assets, Quasi-Monte Carlo (QMC) simulation is combined with a $t$-copula based dependence structure for reliability estimation under uncertainty.

Furthermore, a PPO-based reinforcement learning framework is developed to dynamically optimize portfolio allocation policies under realistic market constraints. The proposed approach is evaluated using a diversified portfolio consisting of major global equity indices across three distinct market regimes, namely the pre-COVID, COVID, and post-COVID periods. The performance of the proposed DRL framework is compared with equal weight benchmark portfolios and the classical NSGA-II multi-objective optimization algorithm.

The major contributions of this work are summarized below:

\begin{enumerate}




\item We develop a reliability based bi-objective portfolio optimization framework for practical portfolio rebalancing under transaction cost constraints using three complementary risk measures: variance, Conditional Value-at-Risk (CVaR), and Entropic Value-at-Risk (EVaR).


\item Reliability constraints are evaluated using Quasi-Monte Carlo (QMC) simulation together with a t-copula dependence model, enabling the framework to capture uncertainty, nonlinear dependence, and tail dependence among global financial assets.


\item We develop a Proximal Policy Optimization (PPO) based deep reinforcement learning framework that learns adaptive portfolio allocation policies under joint risk and reliability constraints, and compare its performance with the classical NSGA-II multi-objective optimization algorithm


\item Extensive experiments are conducted across pre-COVID, COVID, and post-COVID market regimes. The proposed approaches are benchmarked against equal-weight portfolios and NSGA-II under variance, CVaR, and EVaR risk measures.

\item To evaluate the scalability of the proposed framework, additional experiments are performed on the FTSE100 universe, showing that both NSGA-II and PPO remain effective in high-dimensional portfolio optimization while exhibiting distinct portfolio allocation characteristics.

\end{enumerate}


The remainder of this paper is organized as follows. Section 2 presents a comprehensive literature survey on recent advances and research trends in portfolio optimization. Section 3 formulates the portfolio optimization problem and introduces the background concepts and risk measures considered in this study. Section 4 describes the proposed bi-objective portfolio optimization models. Section 5 presents the proposed methodology, including the PPO-based deep reinforcement learning framework. Section 6 details the experimental setup, data preprocessing procedures, and implementation settings. Section 7 reports the experimental results and discussion, including a comparative analysis across different market regimes. Finally, Section 8 concludes the paper and highlights potential directions for future research.

\section{Literature Survey}
Portfolio optimization has been one of the central problems in financial decision making since the seminal work of \cite{markowitz1952portfolio}, who introduced the mean--variance framework to balance expected return and risk. Over the years, several extensions of the classical portfolio optimization problem have been proposed, including multi-objective formulations, stochastic programming approaches, and portfolio models incorporating transaction costs and practical investment constraints \cite{kolm2014portfolio, meghwani2018multi}. Although these traditional optimization frameworks are mathematically rigorous, they are typically based on static assumptions and often struggle to capture the highly dynamic and uncertain nature of real financial markets. To overcome these limitations, researchers increasingly explored machine learning (ML) and deep learning (DL) methods for financial forecasting and portfolio management. Early studies primarily focused on predicting stock returns, volatility, or market trends using neural networks and deep architectures \cite{fischer2018deep, behera2023prediction}. However, in many of these approaches, prediction and portfolio allocation are treated as separate tasks. Specifically, the models first estimate future returns or volatility and subsequently perform portfolio optimization, which may lead to suboptimal decision-making under rapidly changing market conditions.

Recently, Deep Reinforcement Learning (DRL) has emerged as a promising framework for portfolio optimization due to its ability to model sequential decision-making under uncertainty. Unlike traditional optimization methods, DRL enables an agent to continuously interact with the market environment and learn adaptive portfolio allocation policies over time. \cite{chau2025continuous} provides a theoretical foundation for reinforcement learning in portfolio optimization by reformulating continuous-time portfolio selection as an entropy-regularized sequential decision-making problem. The proposed framework demonstrates the feasibility of integrating reinforcement learning with stochastic optimal control under realistic portfolio constraints, thereby motivating subsequent DRL-based portfolio optimization research. \cite{chakraborty2019} investigated the use of DRL algorithms for generating profitable trading strategies within a Markov Decision Process (MDP) framework. Li et al.~\cite{li2019application} proposed a deep reinforcement learning ensemble framework combining PPO, A2C, and DDPG for stock trading. Their method achieved improved risk adjusted returns and outperformed traditional portfolio allocation strategies. Similarly, \cite{gao2020application} proposed a DQN-based portfolio management framework with discretized portfolio weights and dueling network architectures to improve trading performance. Hierarchical DRL architectures incorporating transaction costs were further explored in \cite{gao2021framework}. More recent studies have focused on improving robustness, scalability, and risk sensitivity in portfolio optimization. \cite{JIANG2024101016} proposed a model free DRL framework for dynamic portfolio allocation in high-dimensional financial markets by integrating transaction costs and investor risk aversion into a mean-variance reward function. \cite{ABOLMAKAREM2023109450} developed predictive multi-period multi-objective portfolio optimization models using deep learning techniques to forecast future market behavior. \cite{ndikum2024advancing} introduced an industry grade DRL framework incorporating sim to real methodologies and realistic trading constraints for robust portfolio optimization across multiple asset classes. Furthermore, \cite{Yan2024} proposed a deep portfolio optimization framework that explicitly incorporates transaction costs and risk-aware reward functions into reinforcement learning-based portfolio management.

Several recent studies have investigated advanced reinforcement learning formulations tailored to financial markets. The first work  in this regard was done by  \cite{jang2023deep}. their work  lies in integrating Modern Portfolio Theory (MPT) with deep reinforcement learning through a multimodal tensor-based framework. The paper combines technical indicators and asset correlation information using Tucker tensor decomposition within a DDPG-based portfolio optimization model, enabling dynamic portfolio allocation that captures both temporal market patterns and cross-asset dependencies. The key contribution of \cite{yu2019model} lies in integrating model-based deep reinforcement learning with portfolio optimization, enabling the agent to learn market dynamics through an internal environment model rather than relying solely on historical interactions. This improves sample efficiency and allows more stable long-term portfolio allocation decisions under changing market conditions.
 For example, \cite{JIANG2024101016} developed a model free deep reinforcement learning framework for dynamic portfolio allocation in high-dimensional financial markets, while \cite{Yan2024} incorporated transaction costs and risk-aware reward functions into deep portfolio optimization. These studies highlight the importance of designing portfolio learning frameworks that can effectively capture market dynamics, asset dependencies, and realistic trading constraints.

Despite the significant progress in DRL-based portfolio optimization, most existing studies primarily focus on return maximization and variance-based risk control, while limited attention has been given to reliability-based optimization and tail-risk-aware decision-making. In parallel, reliability-based portfolio optimization frameworks have been studied in the operations research literature. In particular, \cite{sengupta2024bi} developed a bi-objective reliability-based portfolio optimization framework under uncertainty. However, their approach relies on static optimization techniques and does not consider sequential portfolio decision-making through reinforcement learning. Moreover, existing DRL based portfolio optimization methods generally employ standard risk measures such as variance or Sharpe ratio and rarely incorporate coherent tail risk measures such as Conditional Value-at-Risk (CVaR) and Entropic Value-at-Risk (EVaR) within a reliability constrained framework. In addition, practical aspects such as transaction costs are often included only as penalty terms in the reward function, without integrating them into probabilistic reliability constraints.

Motivated by these gaps, this paper proposes a novel framework termed MORP-DRL for multi-objective reliability based portfolio optimization using deep reinforcement learning. The proposed framework integrates reliability constraints, transaction costs, and multiple risk measures including variance, CVaR, and EVaR within a unified DRL based sequential decision making framework. Unlike existing static reliability based optimization methods and conventional DRL portfolio models, the proposed approach simultaneously addresses tail risk control, probabilistic reliability guarantees, and dynamic portfolio rebalancing under realistic market conditions.

\section{Problem Formulation}

\subsection{Risk Measures}

Risk management plays an important role in portfolio optimization under uncertain and volatile market environments. In this study, we consider three commonly used downside risk measures: Value-at-Risk (VaR), Conditional Value-at-Risk (CVaR), and Entropic Value-at-Risk (EVaR), which capture different aspects of tail risk.


For a portfolio return $R_p$ and confidence level $\alpha$, \textbf{VaR} estimates the loss threshold exceeded with probability $1-\alpha$: $\text{VaR}_{\alpha} = \inf\{l : P(R_p \le -l) \ge 1-\alpha\}$.

\textbf{CVaR} extends VaR by measuring the expected loss beyond the VaR threshold: $\text{CVaR}_{\alpha} = E[-R_p \mid -R_p \ge \text{VaR}_\alpha]$.
Unlike VaR, CVaR satisfies coherence properties and captures tail losses more effectively.


 \textbf{EVaR} provides an exponential upper bound on tail risk: $\text{EVaR}_{\alpha}(X) = \inf_{z>0} \left\{ \frac{1}{z} \ln \left( \frac{E[e^{zX}]}{\alpha} \right) \right\}$.
Owing to its strong theoretical properties and sensitivity to extreme downside risk, EVaR serves as an effective alternative risk measure for reliability based and robust portfolio optimization frameworks.
Both CVaR and EVaR are coherent risk measures satisfying the following properties, where \(\rho_{\alpha}(\cdot)\) denotes a generic coherent risk measure at confidence level \(\alpha\):

\[
\begin{aligned}
\text{Monotonicity:}\quad
&X \le Y \;\Rightarrow\; \rho_{\alpha}(X)\le \rho_{\alpha}(Y),\\
\text{Translation Invariance:}\quad
&\rho_{\alpha}(X+c)=\rho_{\alpha}(X)+c,\\
\text{Positive Homogeneity:}\quad
&\rho_{\alpha}(\lambda X)=\lambda\,\rho_{\alpha}(X),\qquad \lambda\ge0,\\
\text{Subadditivity:}\quad
&\rho_{\alpha}(X+Y)\le \rho_{\alpha}(X)+\rho_{\alpha}(Y).
\end{aligned}
\]
 These risk measures are subsequently incorporated into the proposed portfolio optimization framework for evaluating downside risk.

\subsection{Reliability Based Portfolio Optimization}

Reliability Based Design Optimization (RBDO) \cite{hu2024reliability,sengupta2023reliability} incorporates uncertainty directly into optimization through probabilistic constraints. Rather than enforcing deterministic feasibility, RBDO seeks solutions that satisfy constraints with a prescribed reliability level under uncertain conditions. In portfolio optimization, uncertainty naturally arises from fluctuating asset returns and market volatility. Therefore, portfolio decisions should remain feasible with high probability rather than under a single deterministic realization. Following the RBDO framework, the portfolio optimization problem can be formulated as

\begin{equation}
\begin{aligned}
\max_{\mathbf{x}}
\quad &
R_p=\sum_{i=1}^{n}E(r_i)x_i
\\
\text{s.t.}
\quad &
P\!\left(g_j(\mathbf{x},r)\ge0\right)\ge\beta_j,
\quad j=1,\dots,J,
\\
&
\sum_{i=1}^{n}x_i=1,
\\
&
x_i^{L}\le x_i\le x_i^{U},
\quad i=1,\dots,n.
\end{aligned}
\tag{3}
\end{equation}

where \(x_i\) denotes the portfolio weight of the \(i^{th}\) asset, \(r_i\) represents the return of the \(i^{th}\) asset, and \(g_j(\mathbf{x},r)\) denotes the \(j^{th}\) portfolio constraint under uncertain market conditions. The parameter \(\beta_j\in(0,1)\) represents the prescribed reliability level, where larger values imply lower probabilities of constraint violation. To incorporate realistic portfolio rebalancing effects, proportional transaction costs based on changes in portfolio allocations are also considered within the return formulation \cite{JANA2009188,CHEN2015125}.

\section{Proposed Bi-Objective Portfolio Optimization Models}

Most existing reliability based multi-objective portfolio optimization (MORBPO) models \cite{sengupta2024bi} ignore transaction costs during portfolio rebalancing, leading to unrealistic trading behavior and overestimated returns. To address this limitation, proportional transaction costs are explicitly incorporated into the proposed framework.  To the best of our knowledge, transaction cost aware reliability based portfolio optimization under variance , CVaR , and EVaR based risk formulations has not been explored in existing literature.
\vspace{0.5em}

The notations used throughout the proposed models are summarized below:

\begin{itemize}
    \item \(N\): Total number of assets considered in the portfolio.
    
    \item \(w_i\): Portfolio weight allocated to the \(i^{th}\) asset, where \(i = 1,2,\dots,N\) and \(0 \leq w_i \leq 1\).
    
    \item \(w_{i,\min}\) and \(w_{i,\max}\): Minimum and maximum allowable investment weights for the \(i^{th}\) asset, respectively.
    
    \item \(w_i^{0}\): Initial portfolio weight of the \(i^{th}\) asset before rebalancing.
    
    \item \(\bar{r}_i\): Expected return of the \(i^{th}\) asset.
    
    \item \(r_{i,t}\): Return of the \(i^{th}\) asset at time period \(t\).
    
    \item \(\hat{\sigma}_{i,j}\): Estimated covariance between the returns of assets \(i\) and \(j\).
    
    \item \(\alpha\): Confidence level associated with the VaR/CVaR/EVaR calculations.
    
    \item \(T\): Number of scenarios or historical observations used for risk estimation.
    
    \item \(\gamma\): Auxiliary variable associated with the VaR threshold in the CVaR and EVaR formulations.
    
    \item \(\beta_1\): Reliability level associated with the portfolio return constraint.
    
    \item \(\beta_2\): Reliability level associated with the portfolio risk constraint.
    
    \item \(r_p^{*}\): Minimum target portfolio return specified by the investor.
    
    \item \(\sigma_p^{2*}\): Maximum acceptable portfolio variance in Model A.
    
    \item \(CVaR^{*}\): Maximum acceptable CVaR threshold in Model B.
    
    \item \(EVaR^{*}\): Maximum acceptable EVaR threshold in Model C.
    
    \item \(k_i\): Proportional transaction cost coefficient associated with buying or selling the \(i^{th}\) asset.
    
    \item \((x)^+ = \max(x,0)\): Positive-part operator used in the CVaR formulation.
\end{itemize}

\vspace{1em}


To capture different aspects of portfolio risk under uncertainty, we develop three reliability based bi-objective portfolio optimization models. Although all three formulations aim to maximize expected portfolio return while minimizing portfolio risk, they differ in the choice of risk measure employed. Specifically, Model A uses portfolio variance as the risk metric, Model B incorporates Conditional Value-at-Risk (CVaR) to capture downside tail risk, and Model C employs Entropic Value-at-Risk (EVaR), which provides a tighter and more conservative characterization of extreme losses.

\paragraph{Model A: Variance-Based Formulation}
Model A extends the bi-objective reliability-based portfolio optimization framework proposed in \cite{sengupta2024bi} by incorporating proportional transaction costs into the return reliability constraint.
 Reliability constraints ensure that the net portfolio return, after accounting for proportional transaction costs, exceeds a target return \(r_p^*\) with confidence level \(\beta_1\), while the portfolio variance remains below a prescribed threshold \(\sigma_p^{2*}\) with confidence level \(\beta_2\). It is further subject to budget and bound constraints.

\vspace{5pt}

\begin{equation}
\begin{aligned}
\max_{\mathbf{w}} \quad &
\sum_{i=1}^{N} w_i \bar r_i,
\qquad
\min_{\mathbf{w}}
\sum_{i=1}^{N}\sum_{j=1}^{N}
w_iw_j\hat{\sigma}_{i,j}
\\
\text{s.t.}\quad
&
\Pr\!\left[
\sum_{i=1}^{N}w_ir_{i,t}
-
\sum_{i=1}^{N}
k_i|w_i-w_i^0|
\ge r_p^*
\right]\ge\beta_1,
\\
&
\Pr\!\left[
\sum_{i=1}^{N}\sum_{j=1}^{N}
w_iw_j\hat{\sigma}_{i,j}
\le \sigma_p^{2*}
\right]\ge\beta_2,
\\
&
\sum_{i=1}^{N}w_i=1,
\\
&
0\le w_{i,\min}\le w_i\le w_{i,\max}\le1,
\quad i=1,\dots,N.
\end{aligned}
\tag{6}
\end{equation}

\paragraph{Model B: CVaR-Based Formulation:}
Unlike variance, which treats gains and losses symmetrically, CVaR focuses on downside risk by measuring expected losses beyond the VaR threshold. This is also in extension to the framework used in \cite{sengupta2024bi} Therefore, Model B maximizes expected return while minimizing CVaR under reliability and transaction cost constraints.

\vspace{5pt}

\begin{equation}
\begin{aligned}
\max_{\mathbf{w}} \quad &
\sum_{i=1}^{N} w_i \bar r_i,
\qquad
\min_{\mathbf{w}}
\left\{
\frac{1}{\alpha T}
\sum_{t=1}^{T}
\left(
\gamma-\sum_{i=1}^{N}w_ir_{i,t}
\right)^+
+\gamma
\right\}
\\
\text{s.t.}\quad
&
\Pr\!\left[
\sum_{i=1}^{N}w_ir_{i,t}
-
\sum_{i=1}^{N}
k_i|w_i-w_i^0|
\ge r_p^*
\right]\ge\beta_1,
\\
&
\Pr\!\left[
\frac{1}{\alpha T}
\sum_{t=1}^{T}
\left(
\gamma-\sum_{i=1}^{N}w_ir_{i,t}
\right)^+
+\gamma
\le CVaR^*
\right]\ge\beta_2,
\\
&
\sum_{i=1}^{N}w_i=1,
\\
&
0\le w_{i,\min}\le w_i\le w_{i,\max}\le1,
\quad i=1,\dots,N.
\end{aligned}
\tag{7}
\end{equation}

\vspace{10pt}

\paragraph{Model C: EVaR-Based Formulation: }

Unlike CVaR, which measures average tail losses, EVaR provides a tighter and more conservative characterization of extreme risk. Therefore, Model C maximizes expected return while minimizing EVaR under reliability and transaction cost constraints and is an extension of the work done in \cite{sengupta2024bi}

\vspace{5pt}

\begin{equation}
\begin{aligned}
\max_{\mathbf{w}} \quad &
\sum_{i=1}^{N} w_i \bar r_i,
\qquad
\min_{\mathbf{w}}
\gamma \ln
\left(
\frac{1}{T}
\sum_{t=1}^{T}
\exp\left(
-\gamma^{-1}
\sum_{i=1}^{N}
w_i r_{i,t}
\right)
\right)
\\
\text{s.t.}\quad
&
\Pr\!\left[
\sum_{i=1}^{N}w_ir_{i,t}
-
\sum_{i=1}^{N}
k_i|w_i-w_i^0|
\ge r_p^*
\right]\ge\beta_1,
\\
&
\Pr\!\left[
\gamma\ln
\left(
\frac{1}{T}
\sum_{t=1}^{T}
\exp\left(
-\gamma^{-1}
\sum_{i=1}^{N}
w_i r_{i,t}
\right)
\right)
\le EVaR_p^*
\right]\ge\beta_2,
\\
&
\sum_{i=1}^{N}w_i=1,
\\
&
0\le w_{i,\min}\le w_i\le w_{i,\max}\le1,
\quad i=1,\dots,N.
\end{aligned}
\tag{8}
\end{equation}

\section{Proposed Methodology}

We propose a reliability aware multi-objective deep reinforcement learning framework (MORP-DRL) for sequential
portfolio optimization under uncertainty. The framework integrates transaction costs, probabilistic reliability constraints, and multiple risk measures within a Proximal Policy Optimization (PPO) \cite{Schulman2017ProximalPO} based Actor-Critic architecture. Unlike static
portfolio optimization models, the proposed framework enables dynamic portfolio rebalancing by continuously adapting allocations according to evolving market conditions.

\subsection{MDP Based Portfolio Optimization Framework}
\label{sec:mdp_framework}
The portfolio optimization problem is formulated as a Markov Decision Process (MDP) represented by
\(
(\mathcal{S},\mathcal{A},\mathcal{P},R,\gamma),
\)
where \(\mathcal{S}\), \(\mathcal{A}\), \(\mathcal{P}\), \(R\), and \(\gamma\) denote the state space, action space, transition dynamics, reward function, and discount factor, respectively.

\paragraph{State Space (\(\mathcal{S}\)).}
The state at time \(t\) is defined as $s_t=[\mathbf{p}_t,\mathbf{I}_t,\mathbf{w}_{t-1},c_t]$, where \(\mathbf{p}_t\) denotes normalized asset prices, \(\mathbf{I}_t\) represents market indicators and statistical features, \(\mathbf{w}_{t-1}\) corresponds to previous portfolio weights, and \(c_t\) denotes available cash. Including \(\mathbf{w}_{t-1}\) enables the agent to account for portfolio rebalancing and transaction costs.

\paragraph{Action Space (\(\mathcal{A}\)).}
The action corresponds to portfolio allocation weights
\[
a_t=\mathbf{w}_t=[w_{1,t},w_{2,t},\dots,w_{N,t}],
\]
subject to
\[
\sum_{i=1}^{N}w_{i,t}=1,
\qquad
w_{i,\min}\le w_{i,t}\le w_{i,\max}.
\]
Since portfolio weights are continuous, PPO is adopted to solve the optimization problem.

\paragraph{Transition Dynamics (\(\mathcal{P}\)).}
The transition dynamics describe the evolution from \(s_t\) to \(s_{t+1}\) after taking action \(a_t\). Since financial market dynamics are unknown, transitions are learned implicitly from historical data and simulated scenarios. Asset returns are modeled using GARCH(1,1), extreme losses are characterized through EVT, and cross-asset dependence is captured using a \(t\)-copula. QMC simulation is then employed to generate training scenarios.

\paragraph{Reward Function (\(R\)).}
The reward jointly captures return improvement, risk-adjusted performance, and reliability satisfaction(1):
\begin{eqnarray}
\label{reward_1}
r_t
=
\Delta S_t+\Delta R_t+\mathcal{R}_t^{\mathrm{rel}},
\end{eqnarray}
where
\[
\Delta S_t=(S_t-S^{\mathrm{base}})\times100,
\qquad
S_t=\frac{R_t}{\mathcal{R}_t},
\]
and \(\mathcal{R}_t\) denotes the selected portfolio risk measure. The return improvement is defined as
\[
\Delta R_t=(R_t-R_t^{\mathrm{base}})\times100,
\]
with portfolio net return
\[
R_t
=
\mathbf{w}_t^\top\boldsymbol{\mu}_t-TC_t,
\]
where \(TC_t\) represents transaction costs. The reliability component is given by
\[
\mathcal{R}_t^{\mathrm{rel}}
=
\Psi(\rho_t^{\mathrm{ret}},\beta_1)
+
\Psi(\rho_t^{\mathrm{risk}},\beta_2),
\]
where \(\rho_t^{\mathrm{ret}}\) and \(\rho_t^{\mathrm{risk}}\) denote the estimated return and risk reliability probabilities, respectively. The reliability function is defined as
\[
\Psi(\rho,\beta)=
\begin{cases}
30(\rho-\beta), & \rho\ge\beta,\\
-100(\beta-\rho), & \rho<\beta.
\end{cases}
\]

\paragraph{Reliability Constraints:}
To account for uncertainty, reliability measures are estimated using simulated market scenarios and incorporated into the reward design. Let \(\{r_t^{(m)}\}_{m=1}^{M}\) denote \(M\) scenarios generated using the proposed GARCH, EVT, \(t\)-copula, and QMC framework. The empirical return reliability is estimated as

\[
\rho_t^{\mathrm{ret}}
=
\hat P_{\text{return}}
=
\frac{1}{M}
\sum_{m=1}^{M}
\mathbf{1}
\left(
\sum_{i=1}^{N}w_ir_{i,t}^{(m)}
-
\sum_{i=1}^{N}k_i|w_i-w_i^0|
\ge r_p^*
\right),
\]

while the empirical risk reliability is computed as

\[
\rho_t^{\mathrm{risk}}
=
\hat P_{\text{risk}}
=
\frac{1}{M}
\sum_{m=1}^{M}
\mathbf{1}
\left(
\mathcal R^{(m)}(w)
\le
\mathcal R^*
\right).
\]

Here, \(\mathcal R^{(m)}(w)\) denotes the scenario-wise portfolio risk measure (variance, CVaR, or EVaR), and \(\mathcal R^*\) is the corresponding risk threshold. These estimated reliabilities are compared against predefined confidence levels:

\[
\rho_t^{\mathrm{ret}}\ge\beta_1,
\qquad
\rho_t^{\mathrm{risk}}\ge\beta_2.
\]

The resulting reliability estimates are incorporated into the reward component \(\mathcal R_t^{\mathrm{rel}}\) through \(\Psi(\cdot)\), thereby encouraging reliability feasible and risk aware portfolio decisions.

\begin{figure}[htbp]
    \centering
    \includegraphics[width=\textwidth]{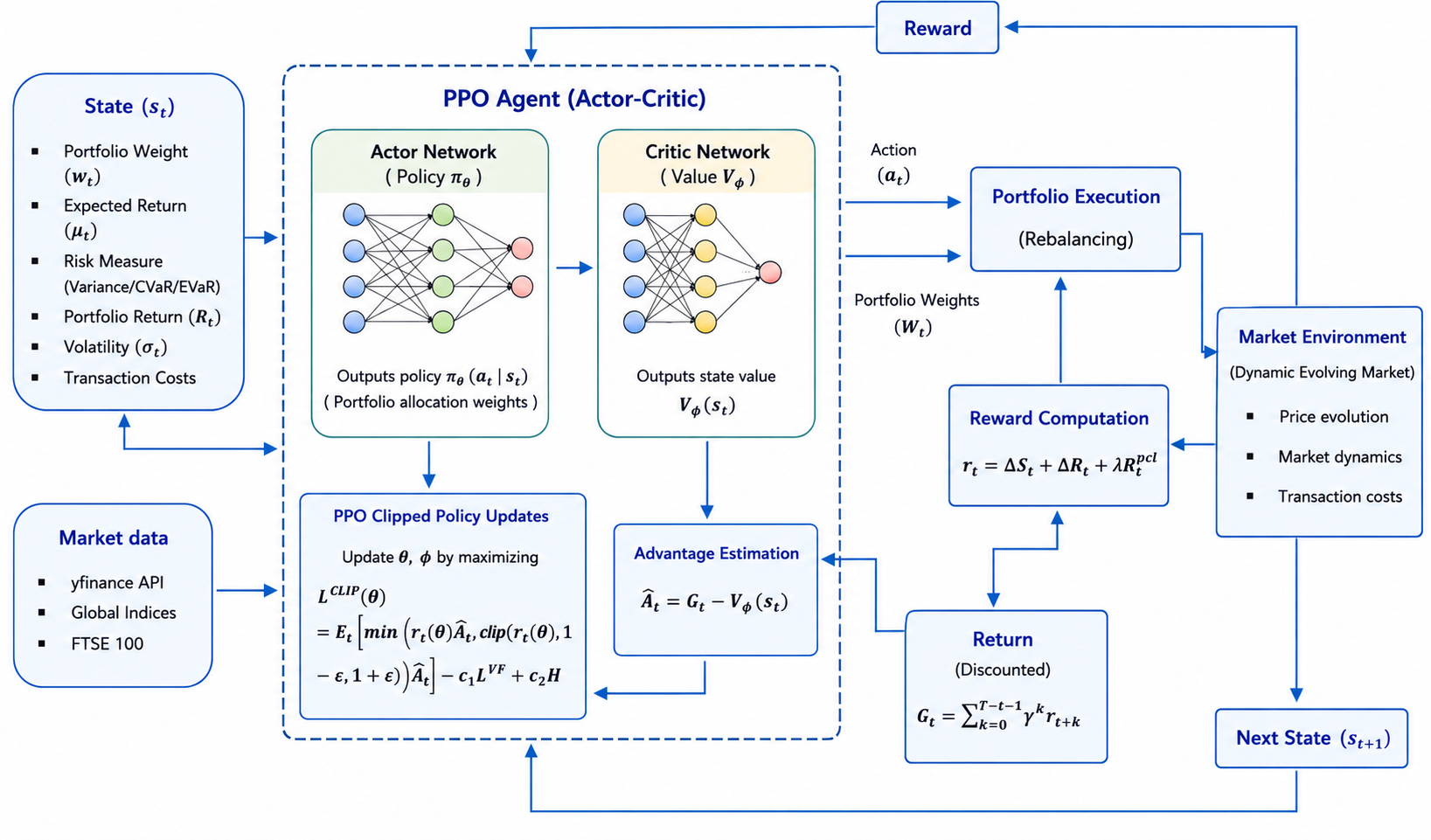}
    \caption{Workflow of the Proposed PPO-Based Portfolio Optimization Model}
    \label{fig:risk_comparison}
\end{figure}

\subsection{Proximal Policy Optimization (PPO) Agent}

The proposed MORP--DRL framework employs a PPO-based Actor--Critic architecture consisting of a policy network (Actor) and a value network (Critic). The Actor generates portfolio allocation decisions, while the Critic estimates the expected cumulative reward of the current market state. The reward signal incorporates portfolio return improvement, risk-adjusted performance, and reliability satisfaction, enabling the agent to learn reliability-aware portfolio strategies. Since portfolio allocation involves continuous decision variables, PPO is adopted due to its effectiveness in continuous action spaces and stable policy updates \cite{Schulman2017ProximalPO}. Policy optimization is performed using the clipped surrogate objective(2):
\begin{equation}
\label{eq:ppo_objective}
L^{\mathrm{CLIP}}(\theta)
=
\mathbb{E}_t
\left[
\min
\left(
r_t(\theta)\hat{A}_t,
\;
\mathrm{clip}\!\left(r_t(\theta),\,1-\epsilon,\,1+\epsilon\right)\hat{A}_t
\right)
\right]
-
c_1 L^{VF}
+
c_2 H
\end{equation}
where
\(
r_t(\theta)
=
\frac{\pi_\theta(a_t|s_t)}
{\pi_{\theta_{\text{old}}}(a_t|s_t)}
\)
denotes the policy probability ratio between the updated and previous policies, and \(\hat A_t\) represents the estimated advantage function. The clipping operation limits excessively large policy updates and improves training stability. Here, \(L^{VF}=\mathbb{E}_t[(V_\phi(s_t)-G_t)^2]\) denotes the value function (critic) loss, which minimizes the prediction error between the estimated state value and the discounted return, while \(H\) denotes the policy entropy that encourages exploration and prevents premature convergence. The coefficients \(c_1\) and \(c_2\) control the relative importance of the value loss and entropy regularization, respectively.  Algorithm~\ref{alg:ppo} outlines the complete training procedure of the proposed MORP–DRL framework. The overall workflow of the proposed PPO-based portfolio optimization model is illustrated in Figure~\ref{fig:risk_comparison}.

\begin{algorithm}[h]
\caption{PPO-Based Reliability-Aware Multi-objective Portfolio Optimization}
\label{alg:ppo}
\small
\begin{algorithmic}[1]

\State Initialize policy network $\pi_\theta(a|s)$ and value network $V_\phi(s)$
\State Initialize PPO hyperparameters $(\eta,\epsilon,\gamma)$
\State Set reliability thresholds $\beta_1,\beta_2$
\State Define portfolio bounds $w_{i,\min},w_{i,\max}$

\For{each training episode}

\State Generate market scenarios using GARCH(1,1), EVT, $t$-copula, and QMC
\State Observe initial state
$
s_0=[\mathbf p_0,\mathbf I_0,\mathbf w_{-1},c_0]
$
\State Initialize trajectory buffer $\mathcal D$

\For{$t=0,1,\dots,T-1$}

\State Sample action:\(
a_t \sim \text{Dirichlet}(\pi_\theta(s_t))
\)

\State Project portfolio weights onto feasible simplex:
\(w_{i,\min}\le w_{i,t}\le w_{i,\max},
\qquad
\sum_{i=1}^N w_{i,t}=1
\)

\State Execute action and observe next state $s_{t+1}$

\State Compute net return $R_t= \mathbf w_t^\top\boldsymbol\mu_t-TC_t$

\State Compute risk measure
$
\mathcal R_t\in
\{\text{Variance,CVaR,EVaR}\}
$

\State Estimate reliability probabilities: $\rho_t^{ret},
\;
\rho_t^{risk}$

\State Compute reward $r_t$ using Eq.~(\ref{reward_1})

\State Store
$(s_t,a_t,r_t,s_{t+1})$
in $\mathcal D$

\State Set $s_t\leftarrow s_{t+1}$

\EndFor

\State Compute discounted cumulative returns:
    \(
    G_t
    =
    \sum_{k=0}^{T-t}
    \gamma^k r_{t+k}
    \)

\State Estimate advantages $  
    \hat{A}_t
    =
    G_t - V_{\phi}(s_t)
    $

\For{each PPO update epoch}

\State Compute policy ratio $
r_t(\theta)
= \frac{\pi_\theta(a_t|s_t)}
{\pi_{\theta_{old}}(a_t|s_t)}
$

\State Update using clipped objective Eq.~(\ref{eq:ppo_objective})

\State Update policy and value networks

\EndFor
\EndFor

\State Return trained policy network $\pi_\theta$

\end{algorithmic}
\end{algorithm}

\section{Experiments}

\subsection{Data Collection and Pre-processing}

The empirical analysis was conducted using a diversified set of major global equity market indices representing different geographic regions and economic environments. The dataset includes the NIFTY 50 (India), S\&P 500 (United States), FTSE 100 (United Kingdom), DAX (Germany), Nikkei 225 (Japan), SSE Composite (China), Hang Seng Index (Hong Kong), KOSPI (South Korea), CAC 40 (France), and ASX 200 (Australia). The inclusion of these indices enables the proposed framework to capture heterogeneous market dynamics and evaluate portfolio behavior under varying economic conditions. In addition to the index-level experiments, a separate large-scale portfolio optimization experiment was conducted using the constituent stocks of the FTSE 100 index in order to assess the scalability and robustness of the proposed methodology in higher-dimensional portfolio settings. Historical daily market data were collected using the \texttt{yfinance} API. The analysis period spans from January 2018 to December 2023 and is divided into three distinct market regimes. The first period, referred to as the \emph{Pre-COVID} regime, covers January 2018 to December 2019 and represents relatively stable market conditions prior to the pandemic. The second period corresponds to the \emph{COVID} regime (January 2020 to December 2021), characterized by higher volatility and significant market disruptions caused by the global pandemic. The final period, denoted as the \emph{Post-COVID} regime, spans January 2022 to December 2023 and captures the subsequent market recovery and stabilization phase.Daily closing prices were transformed into logarithmic returns to ensure time-additive return aggregation. The log-return of asset \(i\) at time \(t\) is computed as
\(
r_{i,t}
=
\ln
\left(
\frac{P_{i,t}}{P_{i,t-1}}
\right),
\)
where \(P_{i,t}\) denotes the closing price of asset \(i\) at time \(t\). 
Financial return series typically exhibit volatility clustering and conditional heteroskedasticity. To model these characteristics, a GARCH(1,1) model was fitted to the return series of each asset. The GARCH framework enables time-varying estimation of conditional volatility and provides a more realistic representation of financial market dynamics. To capture extreme market movements and tail-risk behavior, Extreme Value Theory (EVT) was employed. In particular, excess losses beyond the 95th percentile threshold were modeled using the Generalized Pareto Distribution (GPD). This approach facilitates more reliable estimation of downside risk measures such as CVaR and EVaR, which are incorporated within the proposed portfolio optimization framework. Expected asset returns were estimated using bootstrap resampling in order to account for sampling uncertainty. For each asset, 1000 bootstrap samples were generated from the historical return distribution, and the mean of the resampled returns was used as the expected return estimate. The resulting values were annualized for consistency with the portfolio evaluation metrics.Portfolio risk was estimated using the empirical covariance matrix of daily returns. To obtain annualized risk estimates, the covariance matrix was scaled by a factor of 252.

\subsection{Experimental Setup}

The proposed framework is evaluated under practical portfolio constraints, where the asset weights satisfy
\[
0.03 \leq w_i \leq 0.35,\qquad
\sum_{i=1}^{N} w_i = 1,
\]
with reliability thresholds fixed at \(\beta_1=\beta_2=0.65\). Portfolio rebalancing incorporates proportional transaction costs of 2 basis points at each trading step based on changes in portfolio weights between consecutive periods. The PPO agent adopts an actor--critic architecture and is trained using the AdamW optimizer with a learning rate of \(3\times10^{-4}\). The discount factor and clipping parameter are set to \(\gamma=0.99\) and \(\epsilon=0.2\), respectively, and the agent is trained for 1000 episodes using the PPO clipped surrogate objective (~\ref{eq:ppo_objective}).

\subsection{Benchmarks and Performance Evaluation}

The proposed PPO framework is compared with two benchmark strategies: (i) an equal-weighted (EW) portfolio, where \(w_i=1/N\) for all assets, and (ii) an NSGA-II optimized portfolio solved under the same return, risk, reliability, transaction cost, budget, and bound constraints. Performance is evaluated across the Pre-COVID, COVID, and Post-COVID market regimes using return, volatility, Sharpe ratio, and the corresponding risk measure (Variance, CVaR, or EVaR). In addition, Pareto frontiers are constructed for all optimization models to analyze the trade-off between expected portfolio return and the selected risk measure.

\subsubsection{Performance Metrics}

The performance of the proposed MORP-DRL framework and benchmark strategies is evaluated using several standard portfolio performance measures. These metrics assess not only profitability, but also the associated risk, downside exposure, and computational efficiency of the portfolio optimization framework under different market regimes.

\paragraph {\textbf{Scalability to FTSE100 Constituents:}}
To ensure scalability in high-dimensional portfolio settings, we modify the original quasi-Monte Carlo (QMC) t-copula scenario generation procedure. The baseline approach constructs a single Sobol sequence over the full joint space of assets and time, resulting in a dimensionality of $N \times T$, which quickly exceeds the practical limits of Sobol sequences ($\approx$ 21,201 dimensions) for large portfolios. To address this, we adopt a time-decomposed QMC strategy, wherein Sobol sequences are generated independently at each time step with dimensionality $N$, and subsequently transformed via a Student-t copula using Cholesky-based correlation embedding. This approach preserves cross-sectional dependence across assets while significantly reducing computational complexity and avoiding high-dimensional degeneration of low-discrepancy sequences. Although this introduces an approximation by relaxing inter-temporal dependence, it enables efficient and stable scenario generation for large-scale portfolios such as FTSE 100, making it well-suited for reinforcement learning-based optimization frameworks.

 
\section{Results and Discussion}

Tables \ref{tab:modelA_variance}-\ref{tab:modelC_evar} together with the corresponding Pareto frontiers provide a detailed comparison between the baseline portfolio, NSGA-II optimization, and PPO-based reinforcement learning optimization under different market regimes and risk measures.

\begin{figure*}[t]
\centering

\begin{subfigure}{0.4\textwidth}
    \centering
    \includegraphics[width=\linewidth]{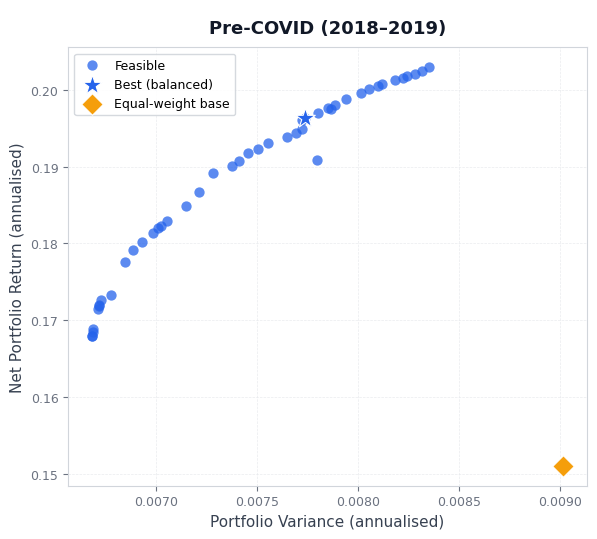}
    \caption{Pre COVID}
    \label{fig:nsga_precovid_var}
\end{subfigure}
\hfill
\begin{subfigure}{0.4\textwidth}
    \centering
    \includegraphics[width=\linewidth]{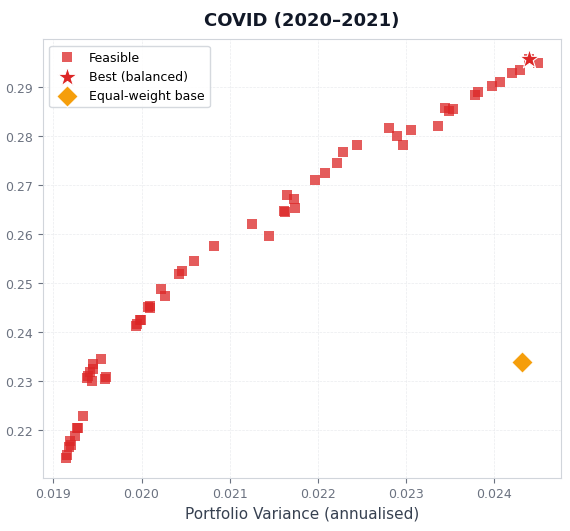}
    \caption{COVID}
    \label{fig:nsga_covid_var}
\end{subfigure}

\vspace{0.5cm}

\begin{subfigure}{0.4\textwidth}
    \centering
    \includegraphics[width=\linewidth]{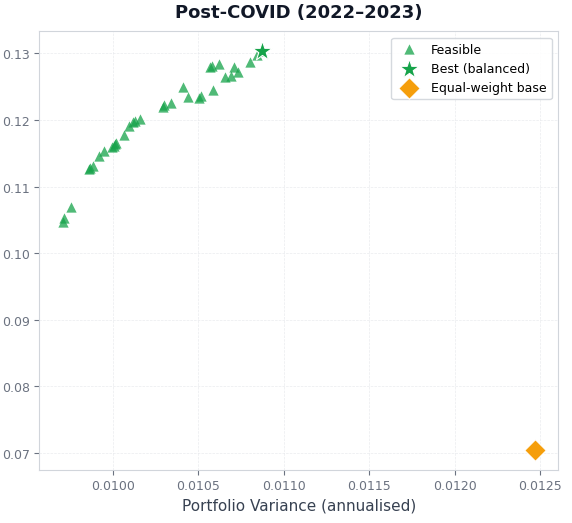}
    \caption{Post COVID}
    \label{fig:nsga_postcovid_var}
\end{subfigure}

\caption{NSGA-II based Pareto frontiers under the Variance risk measure for (a) Pre COVID, (b) COVID, and (c) Post COVID periods.}
\label{fig:nsga_var}
\end{figure*}

\begin{figure*}[t]
\centering

\begin{subfigure}{0.48\textwidth}
    \centering
    \includegraphics[width=\linewidth]{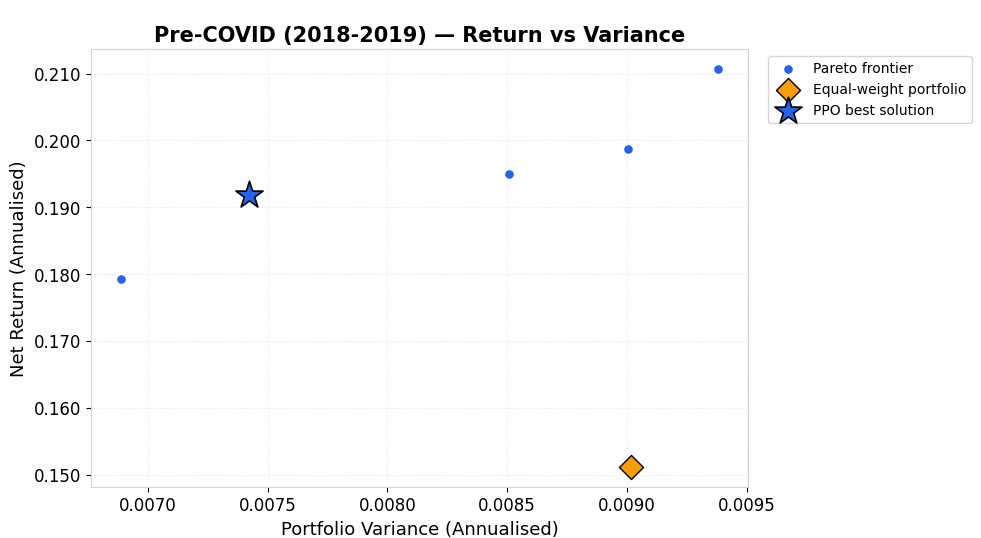}
    \caption{Pre COVID}
    \label{fig:ppo_precovid_var}
\end{subfigure}
\hfill
\begin{subfigure}{0.48\textwidth}
    \centering
    \includegraphics[width=\linewidth]{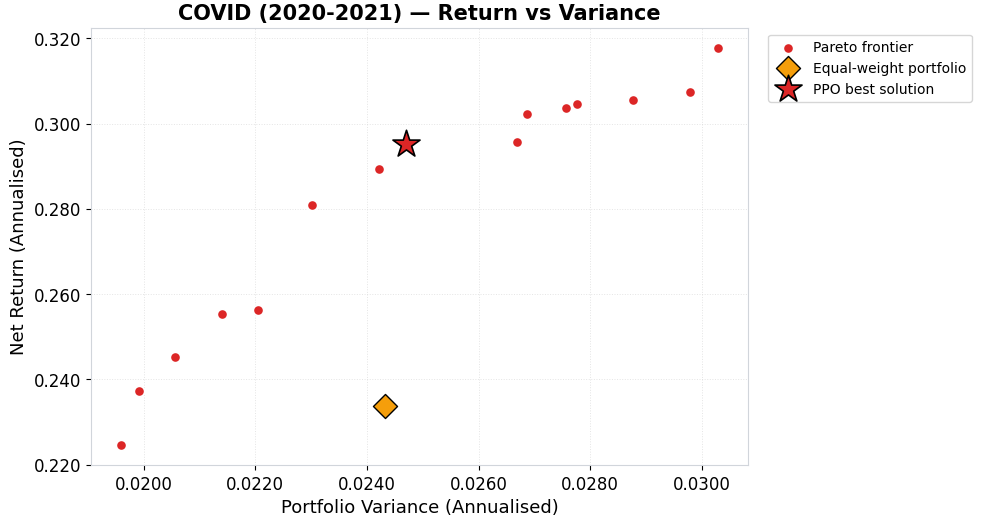}
    \caption{COVID}
    \label{fig:ppo_covid_var}
\end{subfigure}

\vspace{0.5cm}

\begin{subfigure}{0.45\textwidth}
    \centering
    \includegraphics[width=\linewidth]{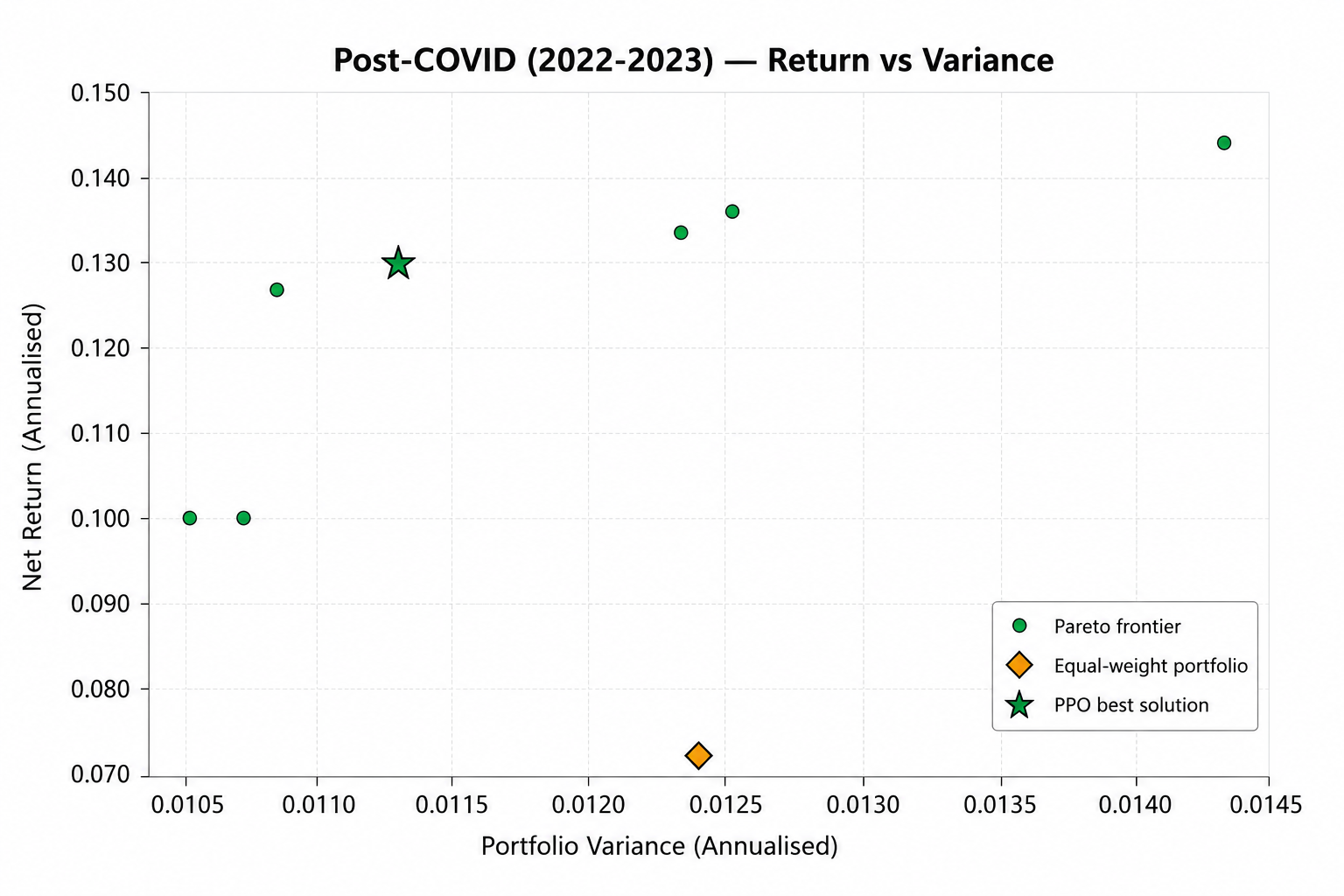}
    \caption{Post COVID}
    \label{fig:ppo_postcovid_var}
\end{subfigure}

\caption{PPO based Pareto frontiers under the Variance risk measure for (a) Pre COVID, (b) COVID, and (c) Post COVID periods.}
\label{fig:PPO_var}
\end{figure*}

\paragraph{Variance-Based Portfolio Optimization (Model A).}
Table \ref{tab:modelA_variance} together with Figures~\ref{fig:nsga_var}--\ref{fig:PPO_var} summarizes the performance of the variance-based portfolio optimization framework across the three market regimes. In all periods, both NSGA-II and PPO produce portfolios that outperform the equal-weight benchmark, demonstrating the effectiveness of optimization based on the mean-variance criterion. During the Pre-COVID period, NSGA-II achieves the highest annualized return (19.63\%) together with the highest Sharpe ratio (2.231), whereas PPO attains slightly lower volatility (8.61\%) and portfolio variance (0.0074). As shown in Figures~\ref{fig:nsga_var} and \ref{fig:PPO_var}, the Pareto frontiers are concentrated within a low-variance region, reflecting the relatively stable market conditions. During the COVID period, the Pareto frontiers shift towards portfolios with substantially higher expected returns accompanied by increased variance, illustrating the elevated market uncertainty. NSGA-II marginally outperforms PPO by achieving the highest return (29.58\%), the highest Sharpe ratio (1.894), and the lowest portfolio variance (0.0244), while PPO provides comparable performance with only a slight increase in risk. In the Post-COVID period, the feasible return--variance region becomes narrower than during the pandemic, indicating a smaller range of attainable risk-return trade-offs. PPO achieves the highest annualized return (13.08\%), whereas NSGA-II maintains lower volatility, lower portfolio variance, and the highest Sharpe ratio (1.250), indicating superior risk adjusted performance. In all three market regimes, the optimized portfolios clearly dominate the equal-weight benchmark in terms of return-risk characteristics. NSGA-II consistently provides stronger risk-adjusted performance under the variance objective, whereas PPO generates competitive portfolios with only marginal differences in return and risk.



\paragraph{CVaR-Based Portfolio Optimization (Model B):}

\begin{figure*}[!t]
\centering

\begin{subfigure}{0.48\textwidth}
    \centering
    \includegraphics[width=\linewidth]{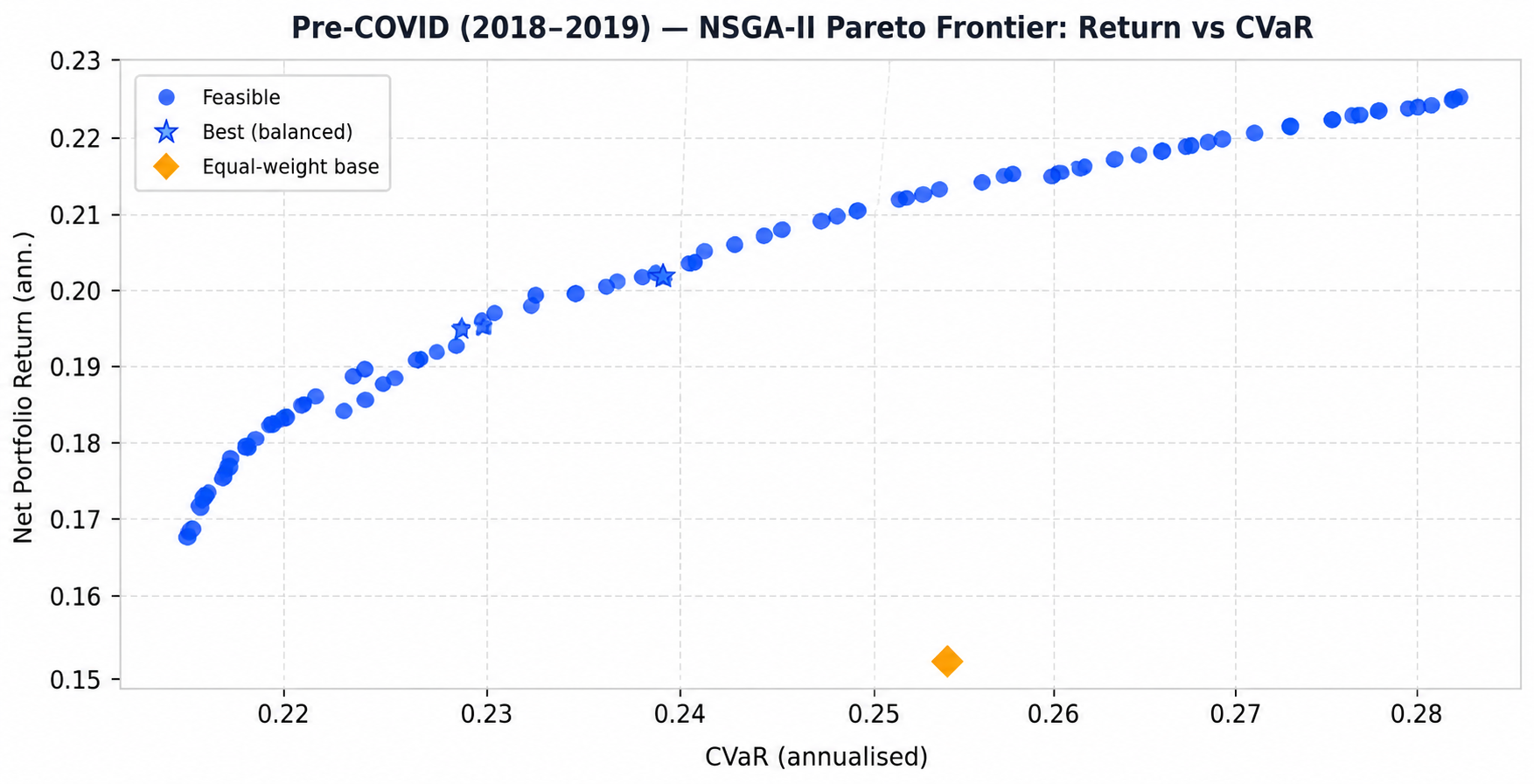}
    \caption{Pre COVID}
    \label{fig:nsga_precovid_cvar}
\end{subfigure}
\hfill
\begin{subfigure}{0.48\textwidth}
    \centering
    \includegraphics[width=\linewidth]{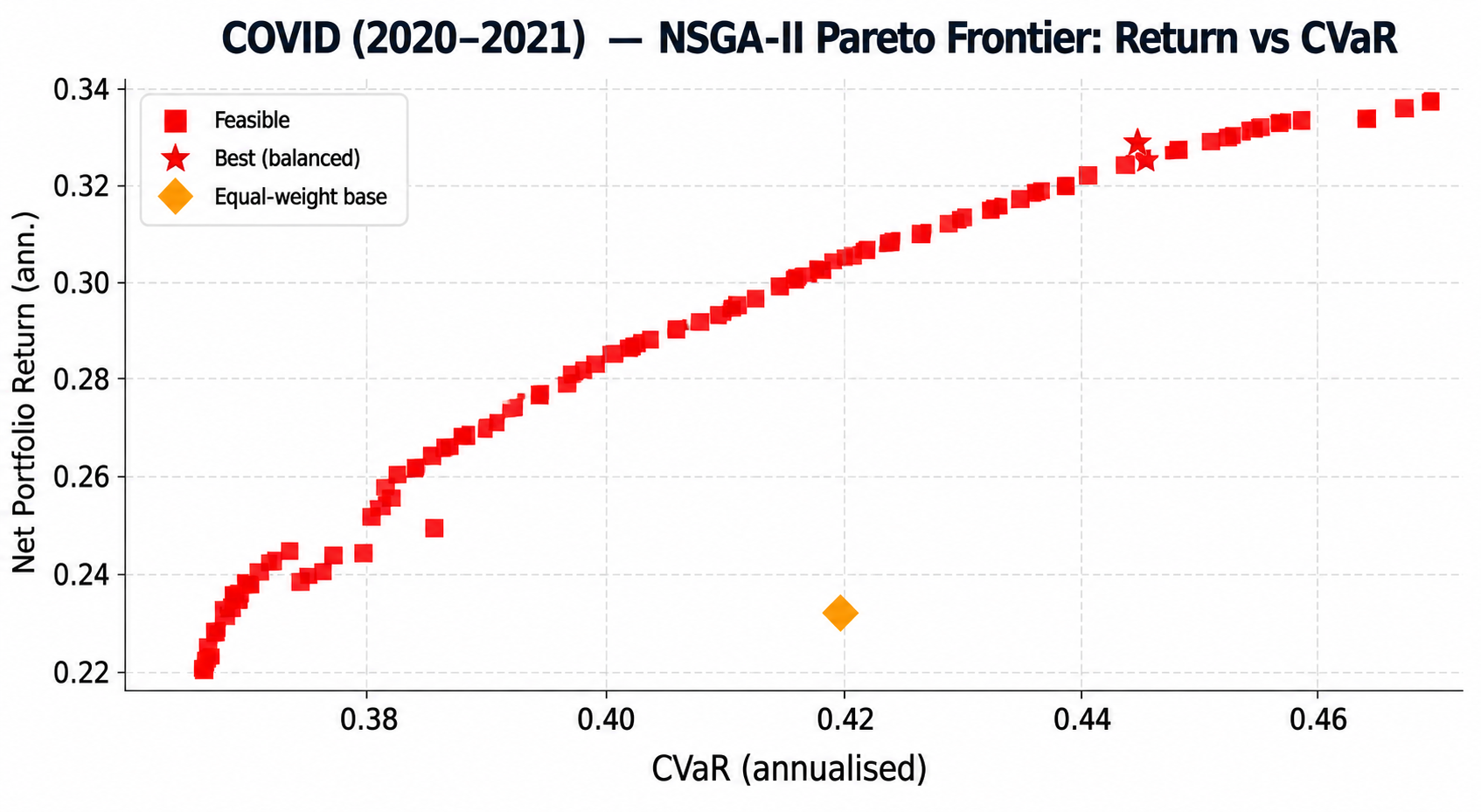}
    \caption{COVID}
    \label{fig:nsga_covid_cvar}
\end{subfigure}

\vspace{0.5cm}

\begin{subfigure}{0.5\textwidth}
    \centering
    \includegraphics[width=\linewidth]{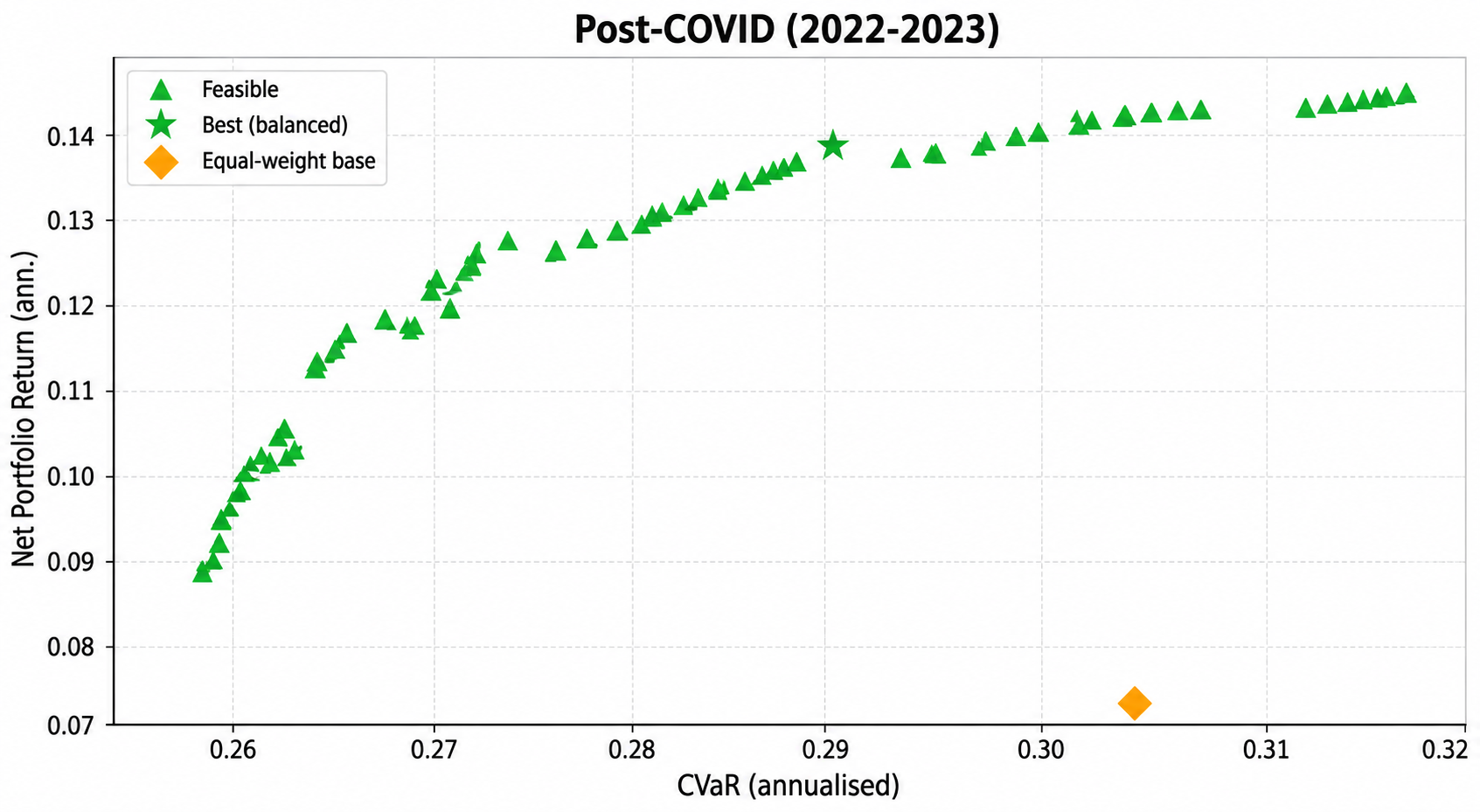}
    \caption{Post COVID}
    \label{fig:nsga_postcovid_cvar}
\end{subfigure}

\caption{NSGA-II based Pareto frontiers under the CVaR risk measure for (a) Pre COVID, (b) COVID, and (c) Post COVID periods.}
\label{fig:nsga_cvar}
\end{figure*}

\begin{figure*}[t]
\centering

\begin{subfigure}{0.48\textwidth}
    \centering
    \includegraphics[width=\linewidth]{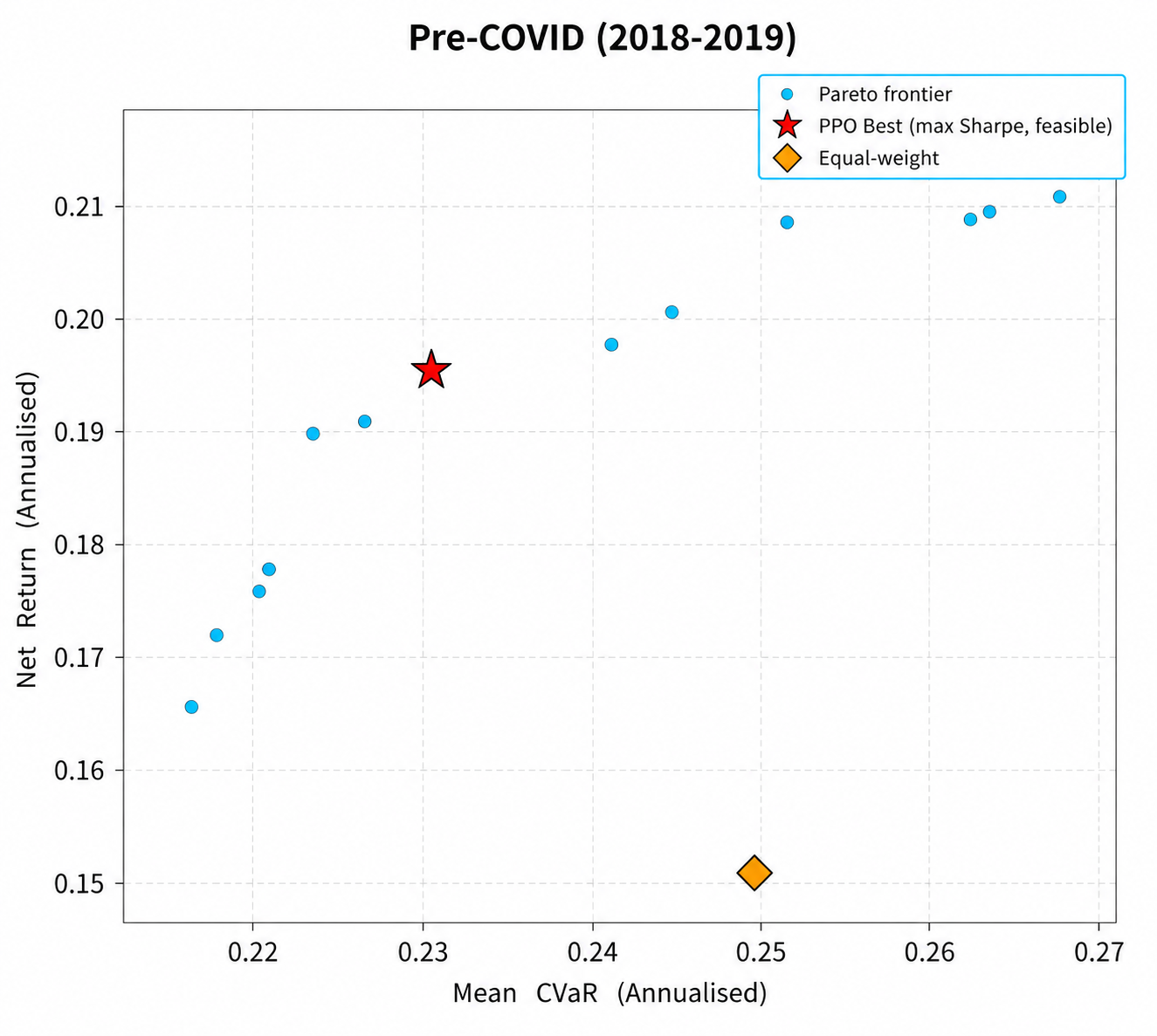}
    \caption{Pre COVID}
    \label{fig:ppo_precovid_cvar}
\end{subfigure}
\hfill
\begin{subfigure}{0.48\textwidth}
    \centering
    \includegraphics[width=\linewidth]{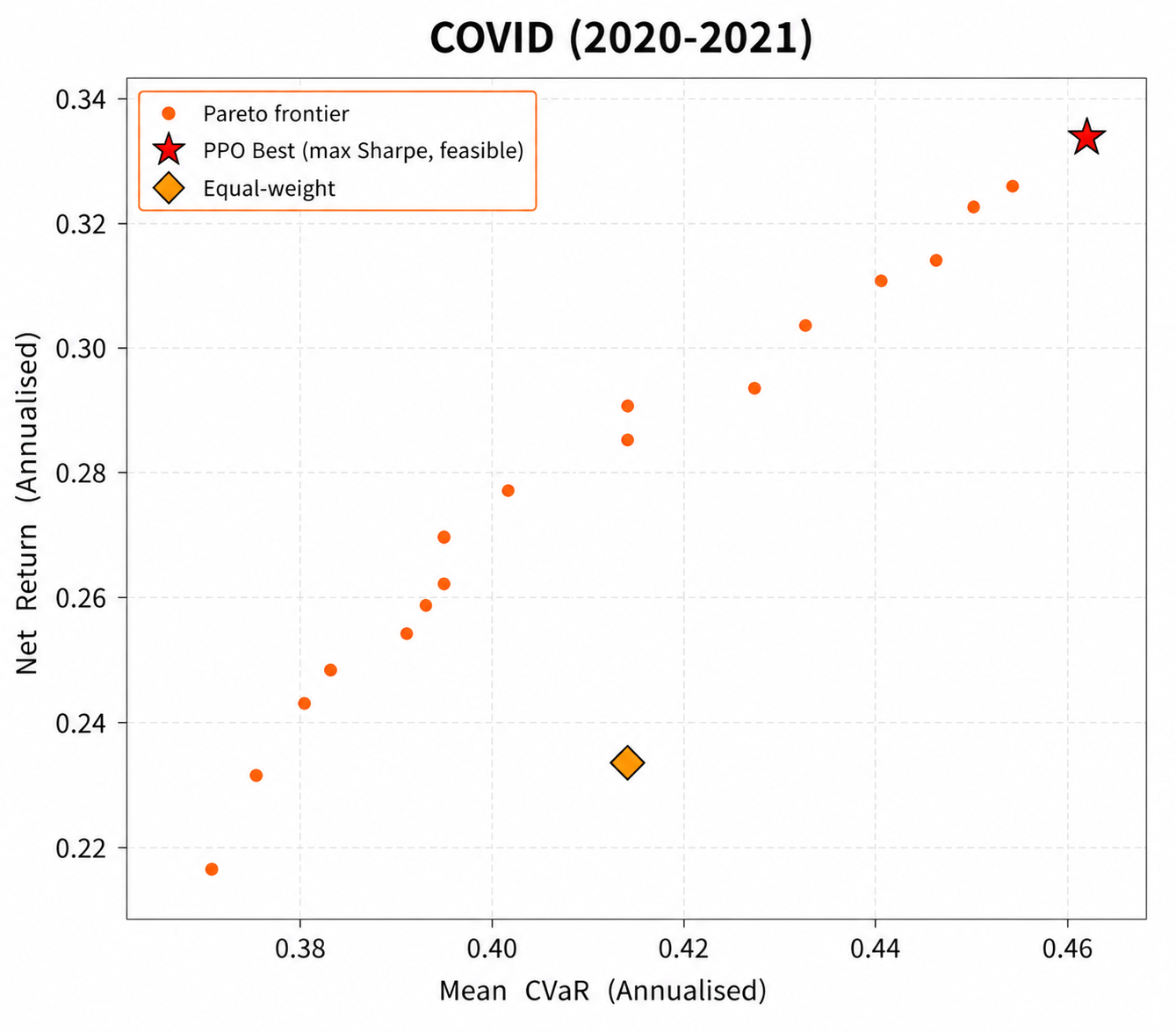}
    \caption{COVID}
    \label{fig:ppo_covid_cvar}
\end{subfigure}

\vspace{0.5cm}

\begin{subfigure}{0.48\textwidth}
    \centering
    \includegraphics[width=\linewidth]{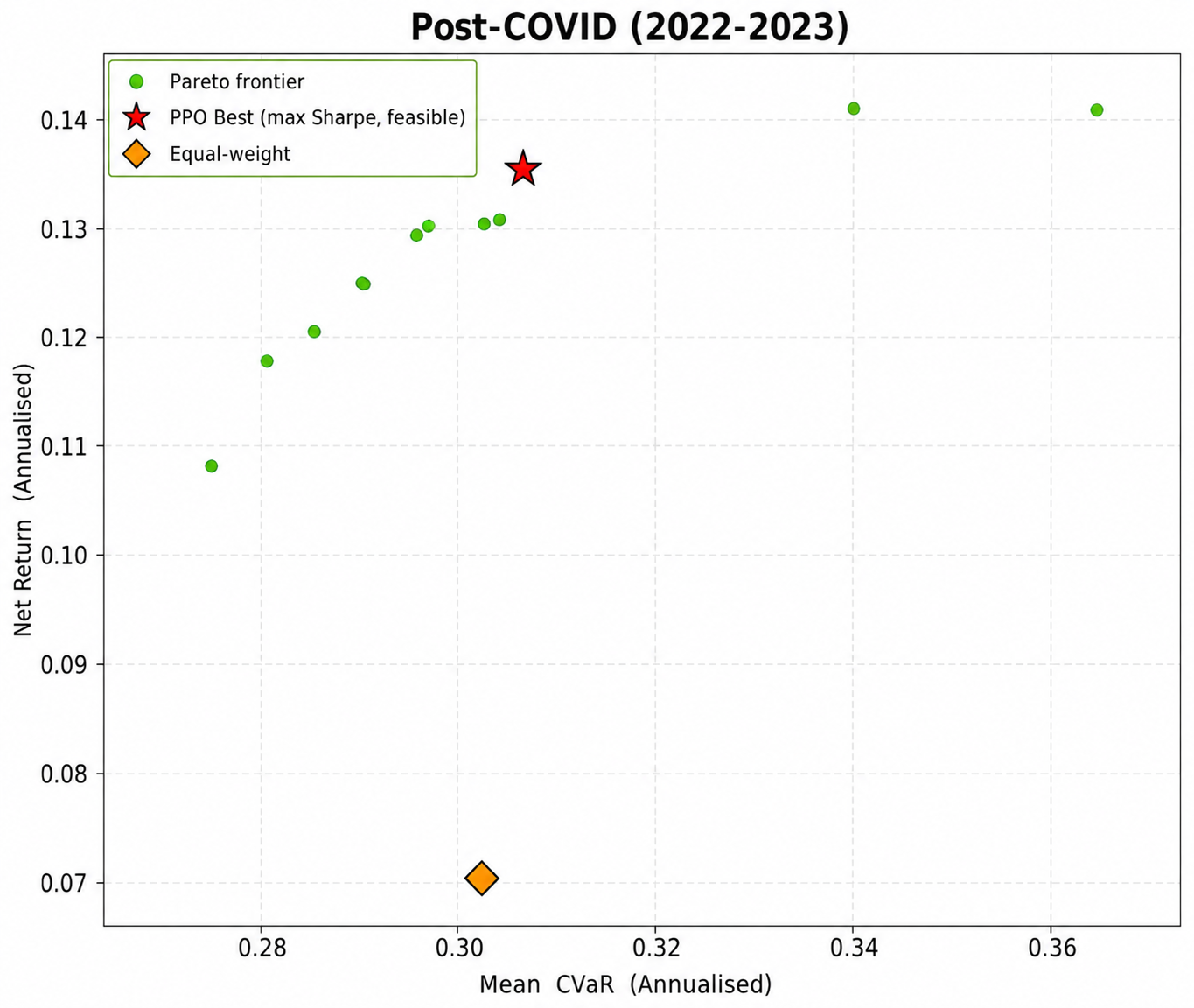}
    \caption{Post COVID}
    \label{fig:ppo_postcovid_cvar}
\end{subfigure}

\caption{Pareto frontiers obtained using PPO under the CVaR risk measure for (a) Pre COVID, (b) COVID, and (c) Post COVID periods.}
\label{fig:ppo_cvar}
\end{figure*}

Table \ref{tab:modelB_cvar} together with Figures \ref{fig:nsga_cvar}--\ref{fig:ppo_cvar} summarizes the performance of the CVaR-based portfolio optimization model, where Conditional Value-at-Risk (CVaR) is used to explicitly control downside tail risk. Across all three market regimes, both NSGA-II and PPO substantially outperform the equal-weight benchmark in terms of return and risk-adjusted performance, demonstrating the effectiveness of incorporating downside-risk considerations into portfolio optimization. During the Pre-COVID period, NSGA-II achieves the highest Sharpe ratio (2.249) while maintaining the lowest CVaR (0.2281), indicating the most favorable balance between return and downside risk. PPO attains a comparable annualized return (19.57\%) but with slightly higher volatility and CVaR. As illustrated in Figures \ref{fig:nsga_precovid_cvar} and \ref{fig:ppo_precovid_cvar}, the Pareto frontiers are concentrated within a relatively narrow return-CVaR region, reflecting the comparatively stable market environment. During the COVID period, the Pareto frontiers shift towards portfolios with substantially higher expected returns accompanied by increased downside risk, illustrating the challenging market conditions. PPO identifies the highest-return portfolio (33.28\%), but this improvement is accompanied by the highest CVaR (0.4601). In contrast, NSGA-II achieves the highest Sharpe ratio (1.914) with a lower CVaR (0.4462), providing a more balanced trade-off between return and downside risk. Moreover, the PPO Pareto frontier spans a wider range of return-CVaR combinations, indicating greater flexibility in generating portfolios corresponding to different levels of downside-risk tolerance. In the Post-COVID period, the Pareto frontiers become more concentrated than during the pandemic, suggesting a narrower range of feasible return-risk trade-offs. NSGA-II marginally outperforms PPO in annualized return (13.88\% versus 13.64\%), Sharpe ratio (1.277 versus 1.241), and CVaR (0.2912 versus 0.2935), although the differences between the two approaches are relatively small. In all three market regimes, both optimization methods produce portfolios that dominate the equal-weight benchmark by achieving superior return-risk characteristics. The results indicate that NSGA-II consistently produces portfolios with lower downside risk and better risk-adjusted performance under the CVaR objective, whereas PPO exhibits greater flexibility by identifying higher return portfolios during periods of elevated market uncertainty, albeit at the expense of increased tail risk.

\paragraph{EVaR-Based Portfolio Optimization (Model C):}

\begin{figure*}[t]
\centering

\begin{subfigure}{0.45\textwidth}
    \centering
    \includegraphics[width=\linewidth]{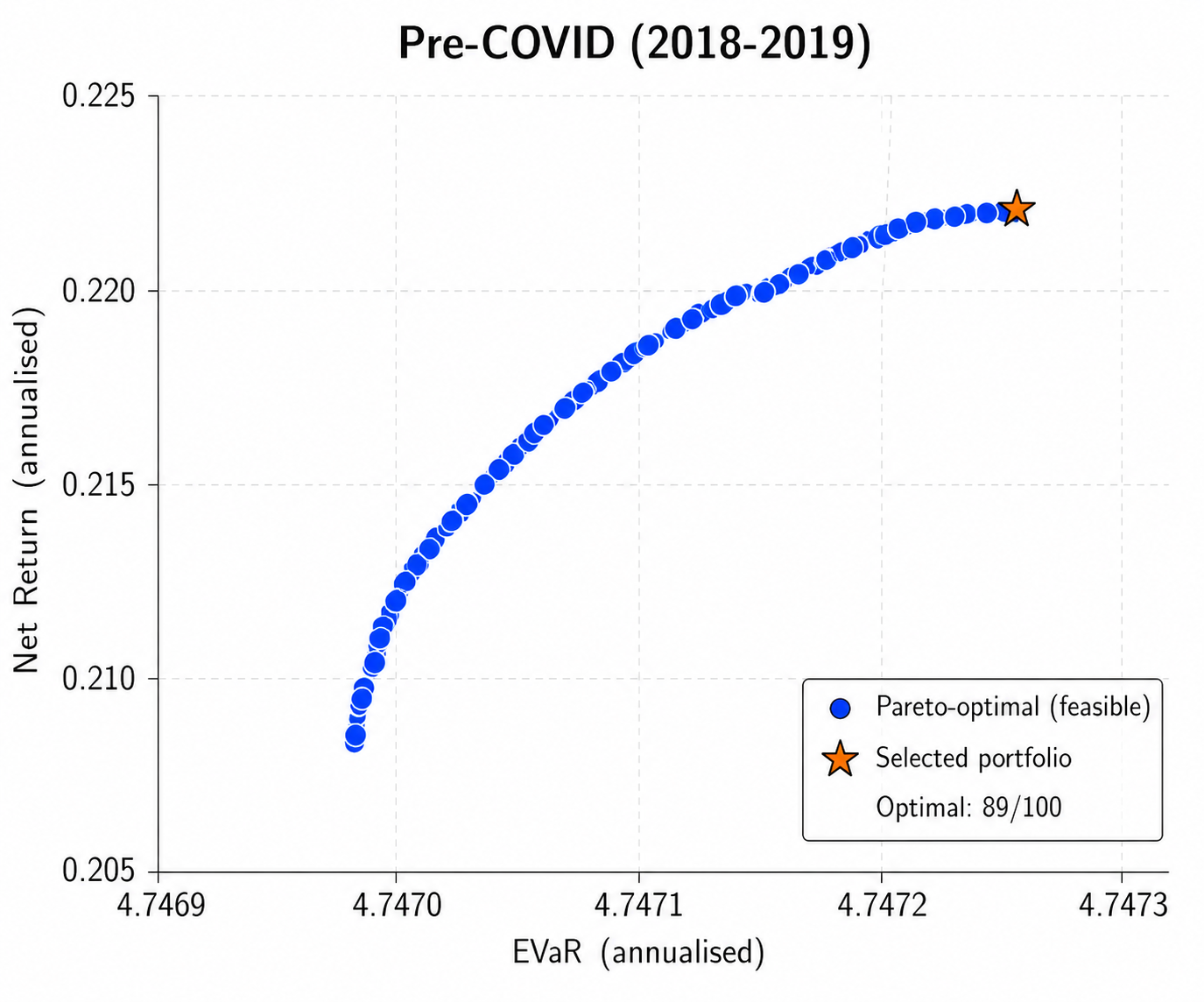}
    \caption{Pre COVID}
    \label{fig:pareto_nsga_precovid_evar}
\end{subfigure}
\hfill
\begin{subfigure}{0.45\textwidth}
    \centering
    \includegraphics[width=\linewidth]{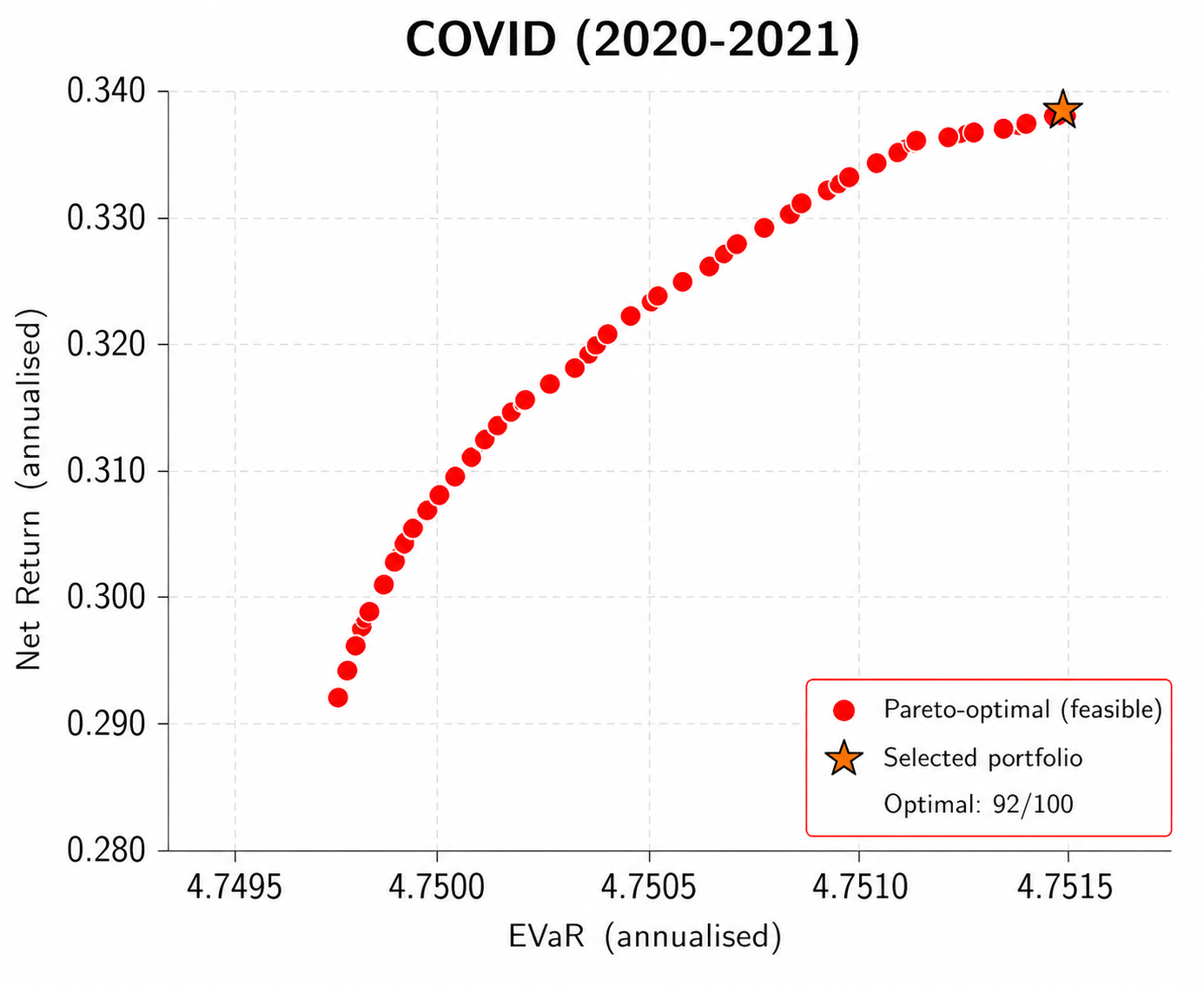}
    \caption{COVID}
    \label{fig:pareto_nsga_covid_evar}
\end{subfigure}

\vspace{0.5cm}

\begin{subfigure}{0.45\textwidth}
    \centering
    \includegraphics[width=\linewidth]{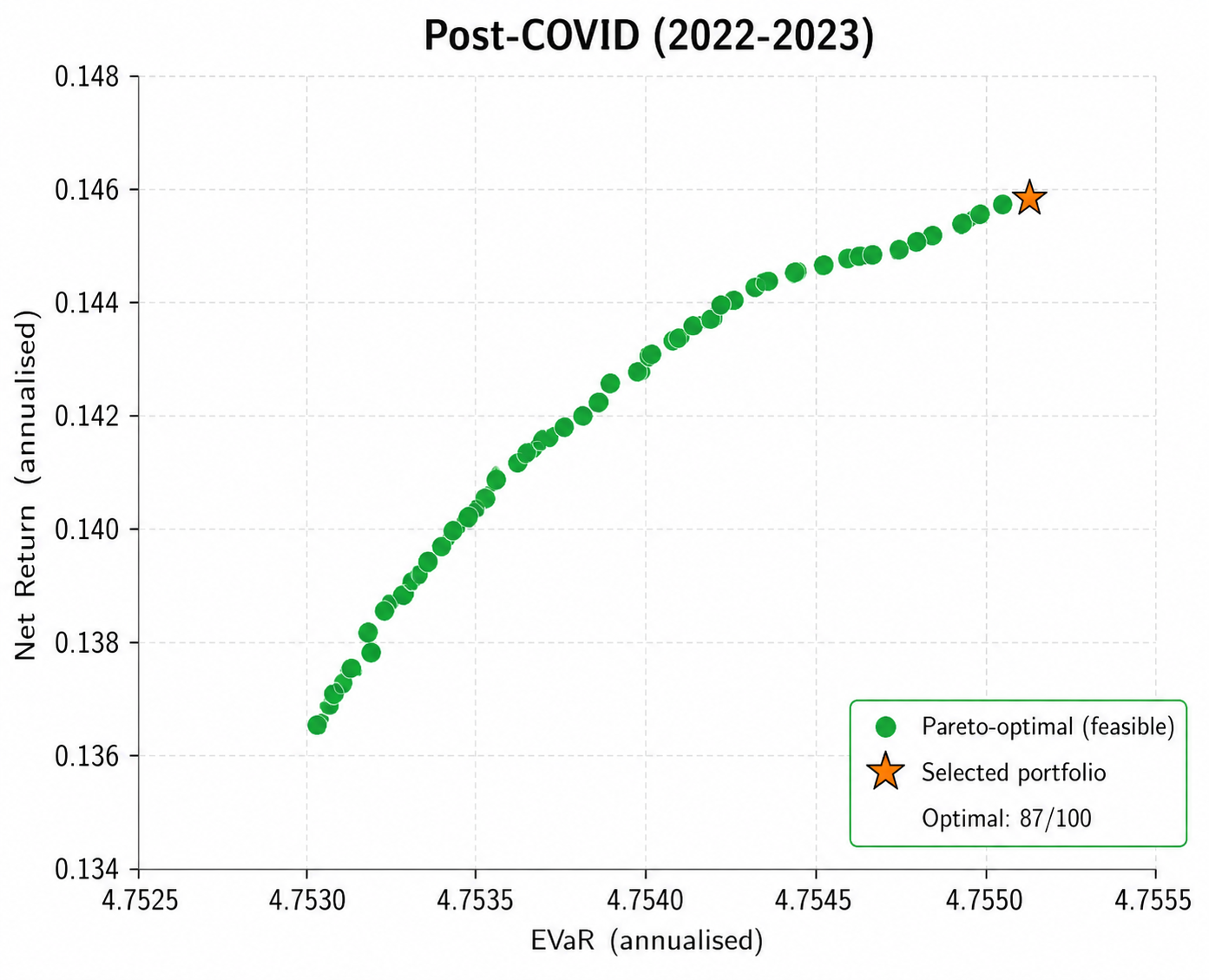}
    \caption{Post COVID}
    \label{fig:pareto_nsga_postcovid_evar}
\end{subfigure}

\caption{Pareto frontiers obtained using NSGA-II under the EVaR risk measure for (a) Pre COVID, (b) COVID, and (c) Post COVID periods.}
\label{fig:pareto_nsga_evar}
\end{figure*}

\begin{figure*}[!t]
\centering

\begin{subfigure}{0.45\textwidth}
    \centering
    \includegraphics[width=\linewidth]{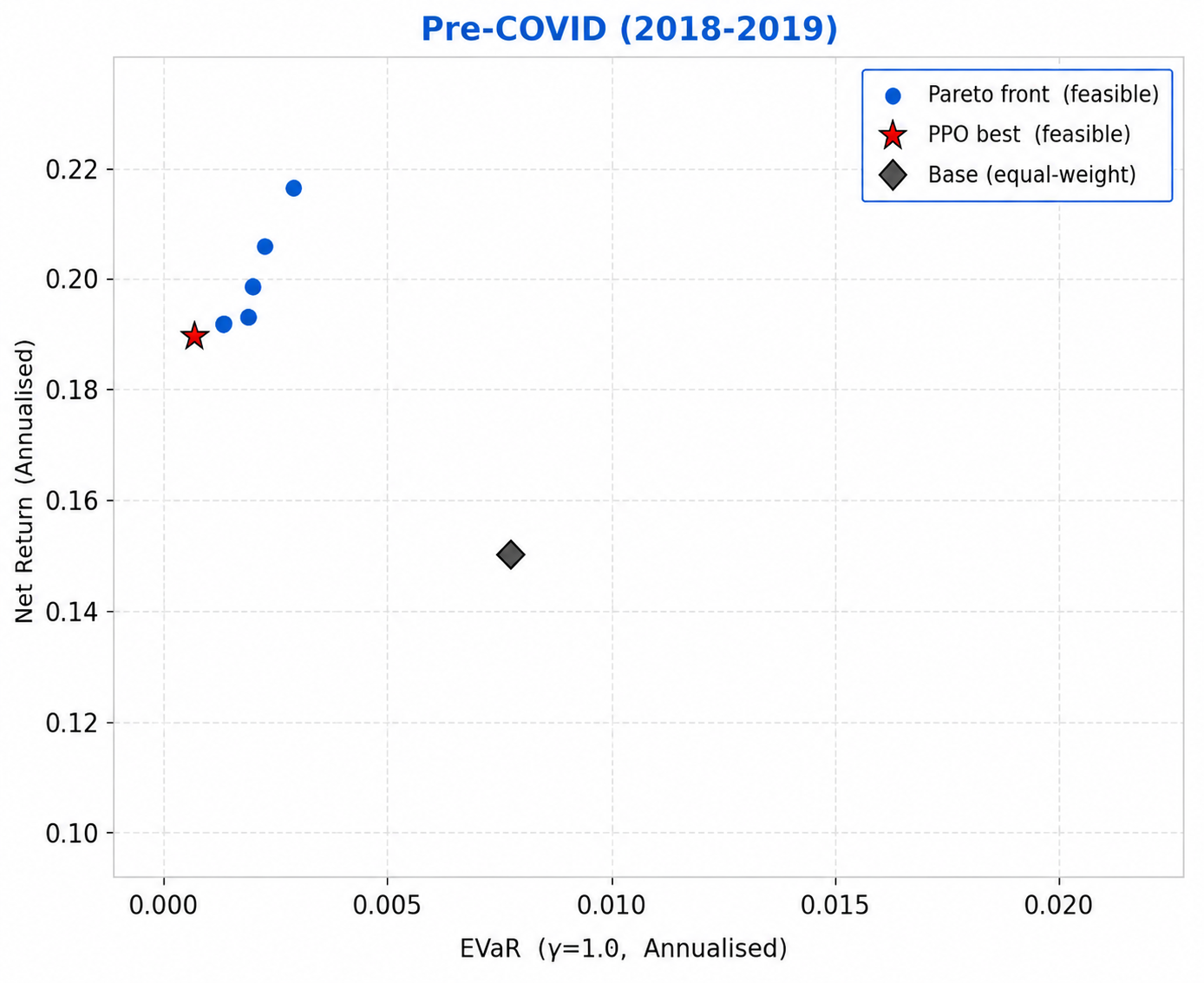}
    \caption{Pre COVID}
    \label{fig:pareto_ppo_precovid_evar}
\end{subfigure}
\hfill
\begin{subfigure}{0.45\textwidth}
    \centering
    \includegraphics[width=\linewidth]{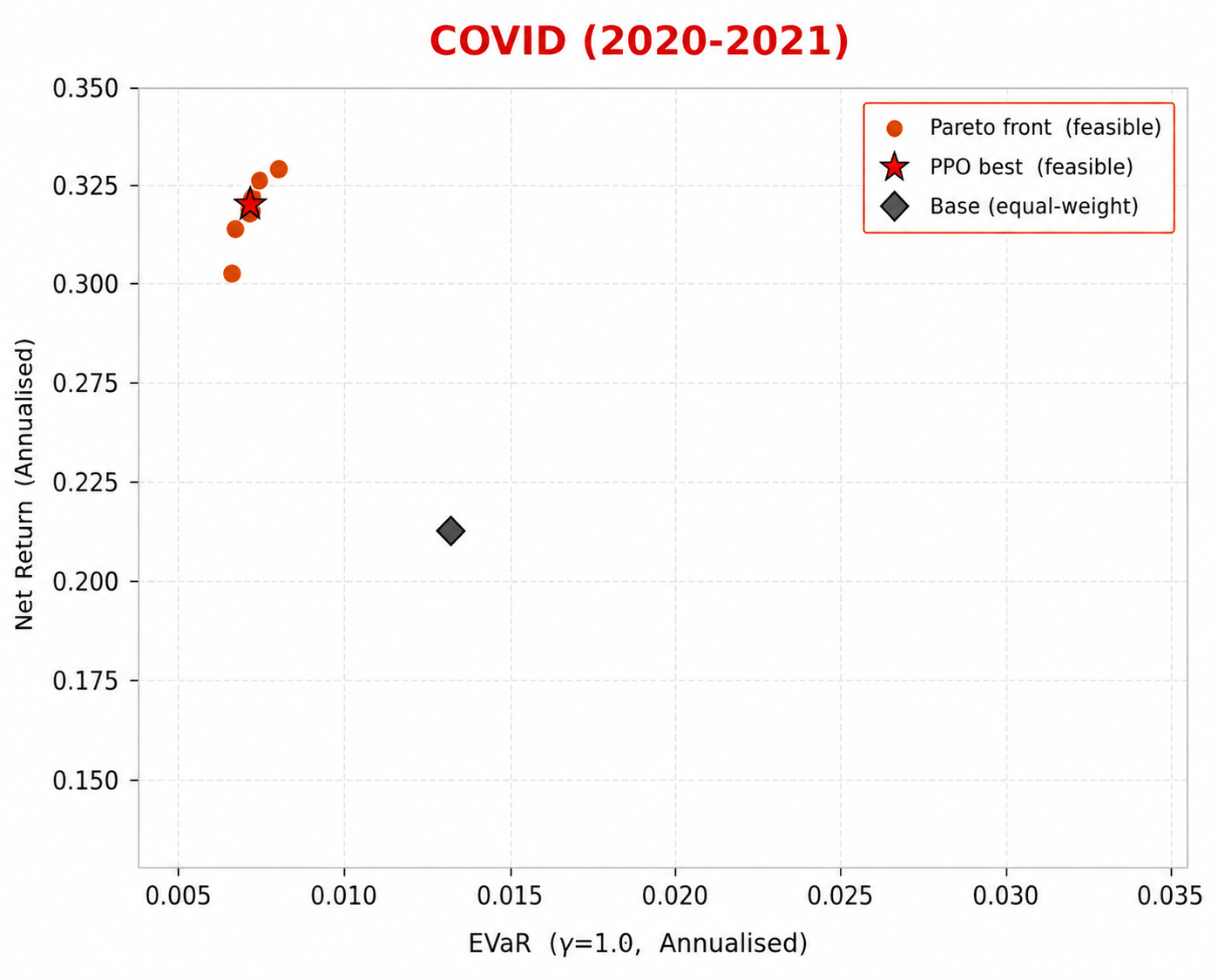}
    \caption{COVID}
    \label{fig:pareto_ppo_covid_evar}
\end{subfigure}

\vspace{0.5cm}

\begin{subfigure}{0.45\textwidth}
    \centering
    \includegraphics[width=\linewidth]{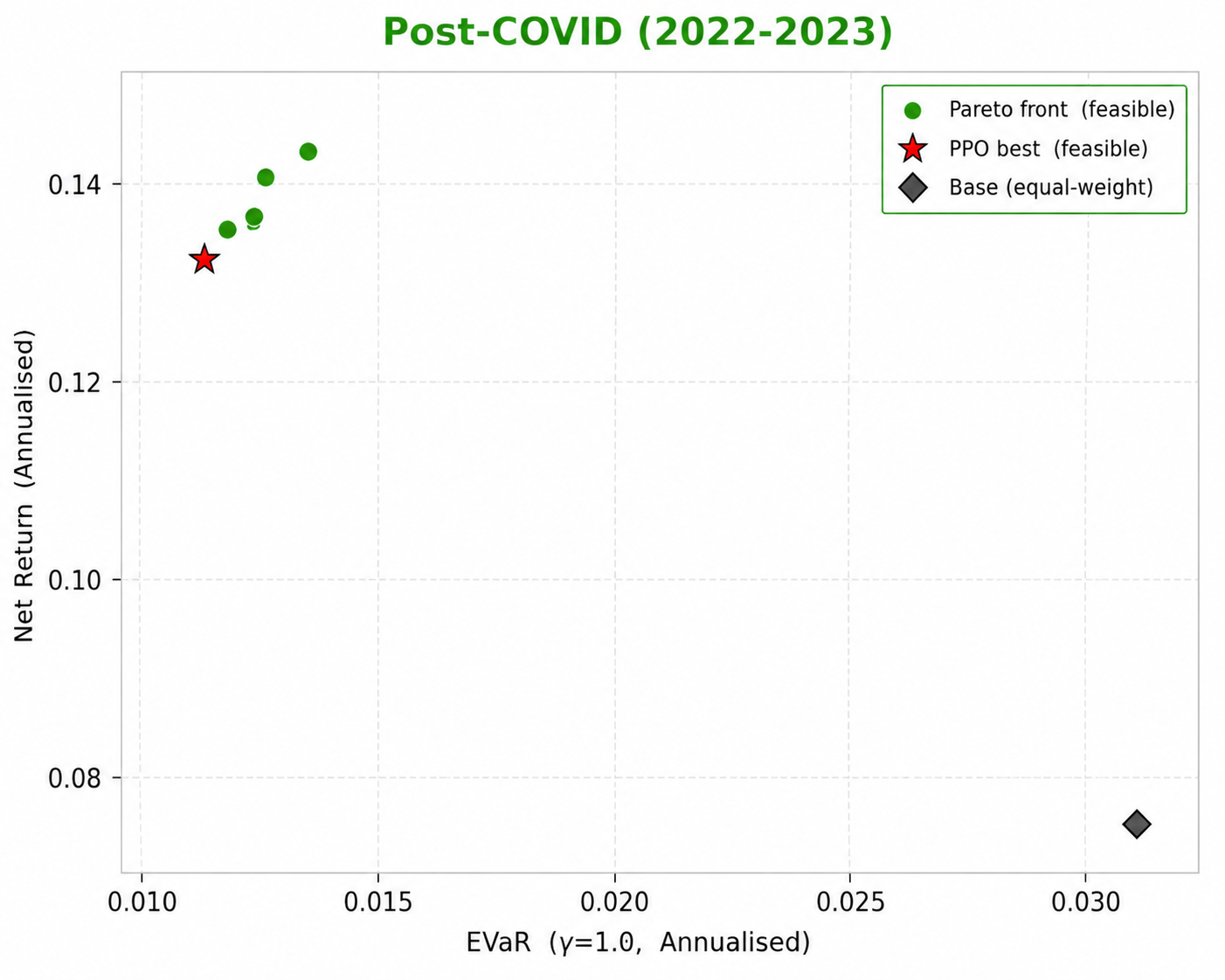}
    \caption{Post COVID}
    \label{fig:pareto_ppo_postcovid_evar}
\end{subfigure}

\caption{Pareto frontiers obtained using PPO under the EVaR risk measure for (a) Pre COVID, (b) COVID, and (c) Post COVID periods.}
\label{fig:pareto_ppo_evar}
\end{figure*}

Table \ref{tab:modelC_evar} together with Figures \ref{fig:pareto_nsga_evar}--\ref{fig:pareto_ppo_evar} presents the performance of the EVaR-based portfolio optimization model, where Entropic Value-at-Risk (EVaR) provides a conservative measure of downside risk by emphasizing extreme loss events. Across all market regimes, both NSGA-II and PPO outperform the equal-weight benchmark, although they exhibit different risk-return characteristics. During the Pre-COVID period, NSGA-II identifies the highest-return portfolio (22.36\%), whereas PPO achieves the highest Sharpe ratio (2.241) by substantially reducing portfolio volatility. Figures \ref{fig:pareto_nsga_precovid_evar} and \ref{fig:pareto_ppo_precovid_evar} show that both methods generate Pareto-efficient portfolios that clearly dominate the equal-weight benchmark. During the COVID period, the Pareto frontiers shift toward portfolios with higher expected returns, reflecting the increased downside risk associated with market uncertainty. NSGA-II achieves the highest annualized return (33.55\%) together with the highest Sharpe ratio (1.906), while PPO produces portfolios with lower volatility, indicating a more conservative risk profile despite a modest reduction in return. The broader range of feasible portfolios obtained by PPO suggests greater flexibility in exploring different return-risk trade-offs under stressed market conditions. In the Post-COVID period, the performance differences between the two methods become smaller. NSGA-II continues to achieve the highest annualized return (14.42\%), whereas PPO maintains lower portfolio volatility and attains the highest Sharpe ratio (1.251), indicating superior risk-adjusted performance. In all three market regimes, both optimization approaches substantially outperform the equal-weight benchmark. The EVaR based results reveal a consistent trade-off between return maximization and conservative risk management. NSGA-II prioritizes higher expected returns, whereas PPO consistently favors lower portfolio volatility and improved risk-adjusted performance.



\paragraph{Comparative Behaviour of NSGA-II and PPO:}
Across the three risk measures and market regimes, NSGA-II and PPO exhibit distinct optimization characteristics. NSGA-II generally achieves stronger risk adjusted performance, consistently requiring substantially lower computational time while producing Pareto-efficient portfolios with competitive or superior returns across most experimental settings. PPO, although computationally more expensive due to repeated policy updates and environment interactions, produces portfolios with performance comparable to NSGA-II and, in several cases, attains either the highest annualized return (e.g., the CVaR-based COVID scenario and the variance-based Post-COVID scenario) or the highest Sharpe ratio (under the EVaR formulation during the Pre-COVID and Post-COVID periods). During periods of elevated market uncertainty, particularly under the EVaR framework, PPO tends to favour portfolios with lower volatility, whereas NSGA-II more frequently identifies portfolios with higher expected returns. Across all experiments, both optimization methods consistently outperform the equal-weight benchmark, demonstrating the effectiveness of multi-objective optimization for portfolio selection. NSGA-II offers a computationally efficient optimization strategy with strong and consistent risk-adjusted performance, while PPO provides a competitive learning-based alternative capable of adapting portfolio decisions through interaction with the market environment.

The differences in optimization behaviour are further illustrated by the portfolio allocation results reported in Appendix A (Tables A1--A3), which present the optimal asset weights under the variance, CVaR, and EVaR formulations across the three market regimes.



\paragraph{FTSE100 Scalability Results:}

The FTSE100 experiments demonstrate that the proposed framework remains effective in high-dimensional portfolio optimization settings. As summarized in Table~\ref{tab:ftse100_summary}, both PPO and NSGA-II consistently outperform the equal-weight benchmark across all three market regimes and risk measures, confirming the scalability of the proposed optimization framework to a larger asset universe. Under the variance formulation, NSGA-II achieves the highest Sharpe ratios in all market regimes, exceeding 2.0 throughout the study, while PPO also provides substantial improvements over the baseline. Similar trends are observed under the CVaR and EVaR formulations, where both optimization methods consistently achieve higher Sharpe ratios than the equal-weight portfolio, with NSGA-II maintaining the strongest risk-adjusted performance across all periods. Although PPO does not surpass NSGA-II in terms of Sharpe ratio, it consistently delivers competitive performance while employing a learning-based optimization strategy. These results demonstrate that both approaches scale effectively to large portfolio optimization problems, with NSGA-II providing superior risk-adjusted performance and computational efficiency, whereas PPO offers a flexible reinforcement learning framework capable of adapting portfolio allocation policies as market conditions evolve.

Additional portfolio concentration statistics are reported in Appendix A (Tables A4, A6, and A7). These results show that NSGA-II generally allocates a larger proportion of portfolio weight to a smaller subset of assets, whereas PPO distributes weights across a larger number of assets, leading to more diversified portfolio allocations. This behaviour is consistent across the variance, CVaR, and EVaR formulations and provides further evidence of the distinct optimization characteristics of the two approaches in high-dimensional portfolio settings.

\begin{table}[h]
\centering
\small

\textbf{Model A} \\[4pt]

\caption{Portfolio Performance Comparison Across Market Regimes --- Variance Risk Measure}
\label{tab:modelA_variance}

\renewcommand{\arraystretch}{1.2}

\begin{tabular}{llcccccc}
\toprule
Period & Method & Return & Volatility & Sharpe & Variance & CPU Time \\
\midrule

\multirow{3}{*}{Pre-COVID}
& Base Portfolio & 0.1511 & 0.0949 & 1.591 & 0.0090 & -- \\
& NSGA-II Optimized & \textbf{0.1963} & 0.0880 & \textbf{2.231} & 0.0077 & \textbf{58.0s} \\
& DRL (PPO) Optimized & 0.1918 & \textbf{0.0861} & 2.226 & \textbf{0.0074} & 848.3s \\

\midrule
\multirow{3}{*}{COVID}
& Base Portfolio & 0.2338 & 0.1559 & 1.499 & 0.0243 & -- \\
& NSGA-II Optimized & \textbf{0.2958} & \textbf{0.1562} & \textbf{1.894} & \textbf{0.0244} & \textbf{57.2s} \\
& DRL (PPO) Optimized & 0.2952 & 0.1572 & 1.878 & 0.0247 & 848.7s \\

\midrule
\multirow{3}{*}{Post-COVID}
& Base Portfolio & 0.0705 & 0.1116 & 0.632 & 0.0125 & -- \\
& NSGA-II Optimized & 0.1303 & \textbf{0.1043} & \textbf{1.250} & \textbf{0.0109} & \textbf{63.0s} \\
& DRL (PPO) Optimized & \textbf{0.1308} & 0.1056 & 1.239 & 0.0112 & 832.4s \\

\bottomrule
\end{tabular}
\end{table}

\begin{table*}[t]
\centering
\small

\textbf{Model B} \\[4pt]

\caption{Portfolio Performance Comparison Across Market Regimes — CVaR Risk Measure}
\label{tab:modelB_cvar}

\renewcommand{\arraystretch}{1.2}

\begin{tabular}{llcccccc}
\toprule
Period & Method & Return & Volatility & Sharpe & CVaR & CPU Time \\
\midrule

\multirow{3}{*}{Pre-COVID}
& Base Portfolio & 0.1511 & 0.0949 & 1.591 & 0.2502 & -- \\
& NSGA-II Optimized & \textbf{0.1958} & \textbf{0.0871} & \textbf{2.249} & \textbf{0.2281} & \textbf{303.4s} \\
& DRL (PPO) Optimized & 0.1957 & 0.0887 & 2.206 & 0.2320 & 725.1s \\

\midrule
\multirow{3}{*}{COVID}
& Base Portfolio & 0.2338 & 0.1559 & 1.499 & 0.4136 & -- \\
& NSGA-II Optimized & 0.3269 & \textbf{0.1708} & \textbf{1.914} & \textbf{0.4462} & \textbf{304.6s} \\
& DRL (PPO) Optimized & \textbf{0.3328} & 0.1754 & 1.897 & 0.4601 & 730.8s \\

\midrule
\multirow{3}{*}{Post-COVID}
& Base Portfolio & 0.0705 & 0.1116 & 0.631 & 0.3028 & -- \\
& NSGA-II Optimized & \textbf{0.1388} & \textbf{0.1087} & \textbf{1.277} & \textbf{0.2912} & \textbf{305.2s} \\
& DRL (PPO) Optimized & 0.1364 & 0.1099 & 1.241 & 0.2935 & 725.8s \\

\bottomrule
\end{tabular}
\end{table*}

\begin{table*}[t]
\centering
\small

\textbf{Model C} \\[4pt]

\caption{Portfolio Performance Comparison Across Market Regimes --- EVaR Risk Measure}
\label{tab:modelC_evar}

\renewcommand{\arraystretch}{1.2}

\begin{tabular}{llcccccc}
\toprule
Period & Method & Return & Volatility & Sharpe & EVaR & CPU Time \\
\midrule

\multirow{3}{*}{Pre-COVID}
& Base Portfolio & 0.1511 & 0.0949 & 1.591 & 4.750728 & -- \\
& NSGA-II Optimized & \textbf{0.2236} & 0.1095 & 2.041 & 4.747798 & \textbf{73.96s} \\
& DRL (PPO) Optimized & 0.1912 & \textbf{0.0853} & \textbf{2.241} & \textbf{0.0013} & 356.8s \\

\midrule
\multirow{3}{*}{COVID}
& Base Portfolio & 0.2338 & 0.1559 & 1.499 & 4.753522 & -- \\
& NSGA-II Optimized & \textbf{0.3355} & 0.1760 & \textbf{1.906} & 4.750670 & \textbf{69.23s} \\
& DRL (PPO) Optimized & 0.3219 & \textbf{0.1705} & 1.888 & \textbf{0.0073} & 351.5s \\

\midrule
\multirow{3}{*}{Post-COVID}
& Base Portfolio & 0.0705 & 0.1116 & 0.631 & 4.757576 & -- \\
& NSGA-II Optimized & \textbf{0.1442} & 0.1230 & 1.172 & 4.754259 & \textbf{67.82s} \\
& DRL (PPO) Optimized & 0.1332 & \textbf{0.1064} & \textbf{1.251} & \textbf{0.0124} & 352.9s \\

\bottomrule
\end{tabular}
\end{table*}

\vspace{0.5em}

\begin{table*}[t]
\small 
\centering
\caption{FTSE100 Performance Comparison Across Risk Measures}
\label{tab:ftse100_summary}
\renewcommand{\arraystretch}{1.2}

\begin{tabular*}{\textwidth}{@{\extracolsep{\fill}} llccc @{}}
\toprule
\textbf{Risk Measure} & \textbf{Period} & \textbf{Baseline} & \textbf{PPO} & \textbf{NSGA-II} \\
\midrule

\multicolumn{5}{c}{\textbf{Variance}} \\
\midrule
& Pre-COVID  & 0.959 & 1.364 & \textbf{2.328} \\
& COVID      & 0.607 & 1.122 & \textbf{2.105} \\
& Post-COVID & 0.645 & 0.971 & \textbf{2.301} \\

\midrule
\multicolumn{5}{c}{\textbf{CVaR}} \\
\midrule
& Pre-COVID  & 0.974 & 1.329 & \textbf{1.926} \\
& COVID      & 0.607 & 1.028 & \textbf{1.202} \\
& Post-COVID & 0.645 & 1.056 & \textbf{1.410} \\

\midrule
\multicolumn{5}{c}{\textbf{EVaR}} \\
\midrule
& Pre-COVID  & 0.959 & 1.535 & \textbf{2.214} \\
& COVID      & 0.607 & 1.057 & \textbf{1.929} \\
& Post-COVID & 0.645 & 1.081 & \textbf{2.211} \\

\bottomrule
\end{tabular*}
\end{table*}







\clearpage

\section{Conclusion and Future Work}

This paper proposes a bi-objective reliability-based portfolio optimization framework in which portfolio allocation is learned as a sequential decision making problem using deep reinforcement learning. The framework jointly optimizes expected return and downside risk while accounting for market uncertainty, transaction costs, and reliability constraints under realistic financial conditions. The empirical study, conducted on ten major global equity indices across the pre-COVID, COVID, and post-COVID periods, demonstrates that the PPO-based policy consistently produces competitive risk-return portfolios and adapts effectively to changing market regimes. The results show that the framework provides improved control of downside and extreme-tail risk under the CVaR and EVaR formulations, leading to more stable portfolio performance during periods of market stress. Pareto frontier analysis further indicates that the learned policies achieve an effective balance between return maximization and risk minimization while satisfying the prescribed reliability constraints. Comparison with NSGA-II shows a clear trade-off between the two optimization approaches. NSGA-II is computationally efficient and often attains higher returns through concentrated portfolio allocations, whereas PPO learns adaptive and more diversified investment strategies that respond effectively to evolving market conditions. The  framework also scales effectively to high-dimensional portfolio optimization, as demonstrated on the FTSE100 constituent stocks using a modified time decomposed quasi-Monte Carlo Student-\(t\) copula  for efficient scenario generation.
Future research may extend the framework to dynamic multi-period portfolio optimization, incorporate additional market frictions such as liquidity and price impact, and investigate hybrid reinforcement learning and evolutionary optimization methods for large-scale portfolio selection. Further improvements may include integrating macroeconomic indicators, news sentiment, and alternative data sources to enable more informed portfolio decisions under rapidly changing market conditions. The use of multi-agent reinforcement learning and distributional reinforcement learning may also provide improved modeling of market interactions and uncertainty. Finally, extending the framework to multi-asset portfolios comprising equities, fixed income securities, commodities, and cryptocurrencies would further enhance its practical applicability for real-world investment management.

\bibliographystyle{unsrt}  
\bibliography{references}  
\clearpage
\FloatBarrier

\appendix
\onecolumn

\setcounter{table}{0}
\renewcommand{\thetable}{A\arabic{table}}


\section{Appendix}

\begin{table}[H]
\centering
\caption{Portfolio Weights Comparison Across Market Regimes -- Variance Optimization}
\label{tab:variance_weights_all}
\renewcommand{\arraystretch}{1.2}

\begin{tabular*}{\textwidth}{@{\extracolsep{\fill}} lcccccc @{}}
\toprule
& \multicolumn{2}{c}{Pre-COVID} & \multicolumn{2}{c}{COVID} & \multicolumn{2}{c}{Post-COVID} \\
\cmidrule(lr){2-3}\cmidrule(lr){4-5}\cmidrule(lr){6-7}
\textbf{Asset} & \textbf{NSGA-II} & \textbf{PPO} & \textbf{NSGA-II} & \textbf{PPO} & \textbf{NSGA-II} & \textbf{PPO} \\
\midrule
SSE      & 3.0  & 6.7  & 25.7 & 7.3  & 3.0  & 3.0  \\
ASX      & 34.5 & 29.6 & 4.3  & 9.7  & 3.1  & 15.1 \\
CAC40    & 9.6  & 17.2 & 3.2  & 7.0  & 3.0  & 3.0  \\
FTSE     & 6.2  & 3.0  & 3.0  & 3.0  & 35.0 & 17.8 \\
DAX      & 3.0  & 3.0  & 3.0  & 3.0  & 3.0  & 13.4 \\
S\&P500  & 10.5 & 14.4 & 25.9 & 28.0 & 11.1 & 5.4  \\
HSI      & 3.0  & 3.4  & 3.0  & 3.0  & 3.0  & 3.0  \\
KOSPI    & 6.8  & 3.0  & 8.2  & 3.9  & 3.0  & 3.0  \\
NIKKEI   & 3.0  & 5.8  & 6.2  & 4.1  & 3.0  & 6.7  \\
NIFTY    & 20.4 & 14.1 & 17.5 & 31.0 & 32.8 & 29.6 \\
\bottomrule
\end{tabular*}
\end{table}
\begin{table}[H]
\centering
\caption{Portfolio Weights Comparison Across Market Regimes -- CVaR Optimization}
\label{tab:cvar_weights_all}
\renewcommand{\arraystretch}{1.2}

\begin{tabular*}{\textwidth}{@{\extracolsep{\fill}} lcccccc @{}}
\toprule
& \multicolumn{2}{c}{Pre-COVID} & \multicolumn{2}{c}{COVID} & \multicolumn{2}{c}{Post-COVID} \\
\cmidrule(lr){2-3}\cmidrule(lr){4-5}\cmidrule(lr){6-7}
\textbf{Asset} & \textbf{NSGA-II} & \textbf{PPO} & \textbf{NSGA-II} & \textbf{PPO} & \textbf{NSGA-II} & \textbf{PPO} \\
\midrule
SSE      & 3.0  & 3.0  & 5.6  & 3.4  & 3.1  & 3.0  \\
ASX      & 25.3 & 27.3 & 3.3  & 3.4  & 4.1  & 10.7 \\
CAC40    & 22.5 & 31.7 & 3.0  & 3.4  & 4.0  & 3.0  \\
FTSE     & 3.1  & 3.0  & 3.0  & 3.4  & 20.8 & 5.2  \\
DAX      & 3.1  & 3.0  & 3.0  & 3.9  & 10.9 & 17.8 \\
S\&P500  & 21.2 & 11.9 & 32.5 & 35.0 & 13.3 & 16.1 \\
HSI      & 3.0  & 3.0  & 3.0  & 3.4  & 3.0  & 3.0  \\
KOSPI    & 3.0  & 5.3  & 9.3  & 3.4  & 3.0  & 3.0  \\
NIKKEI   & 3.0  & 3.4  & 3.4  & 5.8  & 3.0  & 4.7  \\
NIFTY    & 12.7 & 8.4  & 33.9 & 35.0 & 35.0 & 33.5 \\
\bottomrule
\end{tabular*}
\end{table}
\begin{table}[H]
\centering
\caption{Portfolio Weights Comparison Across Market Regimes -- EVaR Optimization}
\label{tab:evar_weights_all}
\renewcommand{\arraystretch}{1.2}

\begin{tabular*}{\textwidth}{@{\extracolsep{\fill}} lcccccc @{}}
\toprule
& \multicolumn{2}{c}{Pre-COVID} & \multicolumn{2}{c}{COVID} & \multicolumn{2}{c}{Post-COVID} \\
\cmidrule(lr){2-3}\cmidrule(lr){4-5}\cmidrule(lr){6-7}
\textbf{Asset} & \textbf{NSGA-II} & \textbf{PPO} & \textbf{NSGA-II} & \textbf{PPO} & \textbf{NSGA-II} & \textbf{PPO} \\
\midrule
SSE      & 3.0  & 3.3  & 3.0  & 3.0  & 3.0  & 3.0  \\
ASX      & 3.0  & 32.7 & 3.0  & 14.4 & 3.0  & 9.7  \\
CAC40    & 33.4 & 17.7 & 4.8  & 10.1 & 19.9 & 5.2  \\
FTSE     & 3.0  & 3.0  & 3.2  & 3.0  & 3.0  & 21.0 \\
DAX      & 11.7 & 3.0  & 3.2  & 3.0  & 34.9 & 6.6  \\
S\&P500  & 33.7 & 21.2 & 35.0 & 28.8 & 3.0  & 17.2 \\
HSI      & 3.1  & 3.0  & 3.1  & 3.0  & 3.0  & 3.0  \\
KOSPI    & 3.0  & 3.0  & 6.7  & 3.0  & 3.0  & 3.0  \\
NIKKEI   & 3.1  & 3.0  & 3.0  & 3.0  & 3.0  & 3.2  \\
NIFTY    & 3.0  & 10.1 & 35.0 & 28.6 & 24.3 & 28.2 \\
\bottomrule
\end{tabular*}
\end{table}
\begin{table}[H]
\centering
\caption{Portfolio Concentration and Weight Distribution(FTSE 100 Variance Risk Measure)}
\label{tab:concentration}
\renewcommand{\arraystretch}{1.2}

\begin{tabular*}{\textwidth}{@{\extracolsep{\fill}} llcccccc @{}}
\toprule
\textbf{Period} & \textbf{Method} & \textbf{Top 10 (\%)} & \textbf{Top 20 (\%)} & \textbf{Top 50 (\%)} & \textbf{$w>1\%$} & \textbf{Max (\%)} & \textbf{Min (\%)} \\
\midrule

\multirow{3}{*}{Pre-COVID}
& PPO      & 36.56 & 55.26 & 88.66 & 38 & 8.54 & 0.10 \\
& NSGA-II  & 64.49 & 86.42 & 95.23 & 22 & 10.00 & 0.10 \\
& Baseline & 10.00 & 20.00 & 50.00 & 97 & 1.03 & 1.03 \\

\midrule
\multirow{3}{*}{COVID}
& PPO      & 43.77 & 64.32 & 90.90 & 32 & 7.88 & 0.10 \\
& NSGA-II  & 66.08 & 84.97 & 94.85 & 20 & 9.89 & 0.10 \\
& Baseline & 10.00 & 20.00 & 50.00 & 99 & 1.01 & 1.01 \\

\midrule
\multirow{3}{*}{Post-COVID}
& PPO      & 39.26 & 59.38 & 89.10 & 31 & 5.45 & 0.10 \\
& NSGA-II  & 69.70 & 86.37 & 94.88 & 19 & 10.02 & 0.10 \\
& Baseline & 10.00 & 20.00 & 50.00 & 99 & 1.01 & 1.01 \\

\bottomrule
\end{tabular*}
\end{table}

\begin{table}[h]
\centering
\caption{Top 10 Holdings (PPO Optimized Portfolio)}
\label{tab:top10_ppo}
\renewcommand{\arraystretch}{1.2}

\begin{tabular*}{\textwidth}{@{\extracolsep{\fill}} l l c l c l c @{}}
\toprule
\textbf{Rank} & \textbf{Pre-COVID} & \textbf{Weight (\%)} & \textbf{COVID} & \textbf{Weight (\%)} & \textbf{Post-COVID} & \textbf{Weight (\%)} \\
\midrule
1 & RTO  & 8.54 & HLMA & 7.88 & UU   & 5.45 \\
2 & 3IN  & 3.72 & FCIT & 5.09 & TSCO & 5.44 \\
3 & DPLM & 3.69 & SMT  & 4.56 & WPP  & 5.02 \\
4 & HWDN & 3.66 & EXPN & 4.42 & WEIR & 3.98 \\
5 & SPX  & 3.56 & NG   & 4.35 & SSE  & 3.85 \\
6 & SGRO & 3.12 & BRBY & 3.89 & AV   & 3.27 \\
7 & HSX  & 2.97 & CNA  & 3.56 & MNDI & 3.26 \\
8 & SHEL & 2.49 & SGRO & 3.52 & CCH  & 3.05 \\
9 & SGE  & 2.48 & AUTO & 3.32 & BA   & 3.04 \\
10& ULVR & 2.34 & GLEN & 3.20 & SMIN & 2.91 \\
\bottomrule
\end{tabular*}
\end{table}

\begin{table}[H]
\centering
\caption{Portfolio Concentration and Weight Distribution (FTSE100 CVaR Risk Measure)}
\label{tab:concentration_cvar}
\renewcommand{\arraystretch}{1.2}

\begin{tabular*}{\textwidth}{@{\extracolsep{\fill}} llcccccc @{}}
\toprule
\textbf{Period} & \textbf{Method} & \textbf{Top 10 (\%)} & \textbf{Top 20 (\%)} & \textbf{Top 50 (\%)} & \textbf{$w>1\%$} & \textbf{Max (\%)} & \textbf{Min (\%)} \\
\midrule

\multirow{3}{*}{Pre-COVID}
& PPO      & 34.99 & 55.75 & 88.21 & 37 & 4.90 & 0.10 \\
& NSGA-II  & 54.93 & 79.06 & 94.41 & 22 & 9.82 & 0.10 \\
& Baseline & 10.00 & 20.00 & 50.00 & 97 & 1.03 & 1.03 \\

\midrule
\multirow{3}{*}{COVID}
& PPO      & 38.79 & 59.77 & 86.35 & 27 & 7.06 & 0.10 \\
& NSGA-II  & 45.24 & 67.19 & 93.25 & 29 & 8.04 & 0.10 \\
& Baseline & 10.00 & 20.00 & 50.00 & 99 & 1.01 & 1.01 \\

\midrule
\multirow{3}{*}{Post-COVID}
& PPO      & 42.71 & 62.11 & 89.78 & 31 & 8.14 & 0.10 \\
& NSGA-II  & 60.61 & 78.99 & 94.39 & 24 & 10.08 & 0.10 \\
& Baseline & 10.00 & 20.00 & 50.00 & 99 & 1.01 & 1.01 \\

\bottomrule
\end{tabular*}
\end{table}

\begin{table}[H]
\centering
\caption{Portfolio Concentration and Diversification Analysis(FTSE100 EVaR Risk Measure)}
\label{tab:diversification}
\renewcommand{\arraystretch}{1.2}

\begin{tabular*}{\textwidth}{@{\extracolsep{\fill}} llcccccc @{}}
\toprule
Period & Method & Top 10 (\%) & Top 20 (\%) & Top 50 (\%) & w$>$1\% & Max (\%) & Min (\%) \\
\midrule

\multirow{3}{*}{Pre-COVID}
& NSGA-II & 57.72 & 77.32 & 94.88 & 26 & 9.80 & 0.10 \\
& PPO     & 42.84 & 63.94 & 91.77 & 31 & 9.65 & 0.10 \\
& Base    & 10.00 & 20.00 & 50.00 & 97 & 1.03 & 1.03 \\

\midrule
\multirow{3}{*}{COVID}
& NSGA-II & 62.27 & 81.44 & 94.30 & 20 & 10.10 & 0.10 \\
& PPO     & 44.58 & 61.53 & 89.36 & 34 & 8.97 & 0.10 \\
& Base    & 10.00 & 20.00 & 50.00 & 99 & 1.01 & 1.01 \\

\midrule
\multirow{3}{*}{Post-COVID}
& NSGA-II & 59.73 & 82.53 & 94.39 & 23 & 10.03 & 0.10 \\
& PPO     & 36.53 & 59.88 & 92.08 & 34 & 5.35 & 0.10 \\
& Base    & 10.00 & 20.00 & 50.00 & 99 & 1.01 & 1.01 \\

\bottomrule
\end{tabular*}
\end{table}

\end{document}